%% file: main.tex
\documentclass[11pt]{article}
\usepackage[preprint]{acl}

\usepackage{times}
\usepackage[T1]{fontenc}
\usepackage[utf8]{inputenc}
\usepackage{microtype}

\usepackage{amsmath}
\usepackage{amssymb}

\usepackage{graphicx}
\usepackage{subcaption}
\usepackage{array}
\usepackage{booktabs}
\usepackage{multirow}
\usepackage{tabularx}
\usepackage[table]{xcolor}
\usepackage{adjustbox}

\usepackage{pifont}
\usepackage{comment}
\usepackage{hyphenat}
\usepackage{kotex}
\usepackage{fontawesome5}
\usepackage{orcidlink}

\newcommand{\xmark}{\ding{55}}
\newcommand{\cmark}{\ding{51}}

\definecolor{LightBlue}{RGB}{221,234,248}
\definecolor{LightRed}{RGB}{255,230,235}
\definecolor{MyBlue}{RGB}{0,112,192}
\definecolor{MyRed}{RGB}{192,0,0}
\definecolor{LightGrayLine}{gray}{0.85}

\makeatletter
\let\ACL@origparagraph\paragraph
\def\paragraph#1{\ACL@origparagraph{#1.}}
\renewcommand\appendix{\par
  \setcounter{section}{0}%
  \setcounter{subsection}{0}%
  \setcounter{subsubsection}{0}%
  \gdef\thesection{Appendix~\@Alph\c@section}%
  \setcounter{equation}{0}%
  \gdef\theequation{\@Alph\c@section.\arabic{equation}}}
\def\blfootnote{\gdef\@thefnmark{}\@footnotetext}
\makeatother

\newcommand{\HFicon}{%
  \IfFileExists{hf-logo.pdf}{\raisebox{-0.16em}{\includegraphics[height=1em]{hf-logo.pdf}}}{%
  \IfFileExists{hf-logo.png}{\raisebox{-0.16em}{\includegraphics[height=1em]{hf-logo.png}}}{%
  \faDatabase}}}
\newcommand{\reslink}[3]{\mbox{{\large#1}\;\href{#2}{\bfseries #3}}}
\newcommand{\ressep}{\hspace{1.1em}\textcolor{black!45}{\rule[-0.15em]{0.5pt}{1em}}\hspace{1.1em}}

\title{Leveraging Fine-grained Error Correction in Korean Speech Recognition for Consultation Services}

\author{
  \textbf{Yonghyun Jun\textsuperscript{1}},
  \textbf{Jimin Lee\textsuperscript{1}},
  \textbf{Hwan Chang\textsuperscript{1}},
  \textbf{Dongho Shin\textsuperscript{2}},
  \textbf{Seolah Kim\textsuperscript{3}},
  \textbf{Hwanhee Lee\textsuperscript{1}}\thanks{\,Corresponding author.}
  \\[2pt]
  \textsuperscript{1}Chung-Ang University \quad
  \textsuperscript{2}Korea Local Information Research \& Development Institute \quad
  \textsuperscript{3}SK intellix
  \\[2pt]
  \texttt{\{zgold5670, ljm1690, hwanchang, hwanheelee\}@cau.ac.kr} \\
  \texttt{dhs@klid.or.kr}, \quad \texttt{seolah.kim@sk.com}
  \\[4pt]
  \reslink{\faGithub}{https://github.com/yonghyun010102/Dasan_Project}{Code}\ressep
  \reslink{\HFicon}{https://huggingface.co/datasets/zgold5670/DasanCallDial}{Dataset}\ressep
  \reslink{\faLink}{https://doi.org/10.1016/j.engappai.2026.116038}{DOI}
}
\begin{document}
\maketitle

\blfootnote{Accepted manuscript of an article published in \emph{Engineering
Applications of Artificial Intelligence} 183 (2026) 116038. Version of record:
\href{https://doi.org/10.1016/j.engappai.2026.116038}{doi:10.1016/j.engappai.2026.116038}.
\copyright{} 2026. This manuscript version is made available under the
CC-BY-NC-ND 4.0 license,
\url{https://creativecommons.org/licenses/by-nc-nd/4.0/}.}
\blfootnote{\textsuperscript{2,3}This work was conducted while the authors were
affiliated with the 120 Dasan Call Foundation.}

\input{0_abstract}

\input{1_introduction}
\input{2_related_work}
\input{3_dataset}

\input{4_methodology}

\input{5_experiments}
\input{6_analyses}
\input{7_discussion}

\input{8_conclusion}

\section*{CRediT authorship contribution statement}
\textbf{Yonghyun Jun:} Writing -- original draft, Visualization, Validation,
Methodology, Investigation, Formal analysis, Conceptualization.
\textbf{Jimin Lee:} Writing -- original draft, Visualization, Methodology,
Conceptualization.
\textbf{Hwan Chang:} Writing -- review \& editing, Writing -- original draft,
Validation, Methodology.
\textbf{Dongho Shin:} Data curation.
\textbf{Seolah Kim:} Data curation.
\textbf{Hwanhee Lee:} Writing -- review \& editing, Supervision, Project
administration, Funding acquisition, Conceptualization.

\section*{Declaration of Generative AI and AI-assisted technologies in the writing process}
During the preparation of this work the author(s) used Claude-Opus-4.8 and
GPT-5.6-Sol in order to improve readability. After using this tool/service,
the author(s) reviewed and edited the content as needed and take(s) full
responsibility for the content of the publication.

\section*{Declaration of competing interest}
The authors declare that they have no known competing financial interests or
personal relationships that could have appeared to influence the work reported
in this paper.

\section*{Acknowledgments}
This work was supported by the Chung-Ang University Graduate Research
Scholarship in 2025. This work was also supported by the Institute of
Information \& Communications Technology Planning \& Evaluation (IITP) grant
funded by the Korea government (MSIT) (No. RS-2021-II211341; Artificial
Intelligence Graduate School Program at Chung-Ang University) and
(No. RS-2026-25546026; Leading Generative AI Human Resources Development).

\section*{Data availability}
The code is available at
\href{https://github.com/yonghyun010102/Dasan_Project}{Github}, and the
dataset is available at
\href{https://huggingface.co/datasets/zgold5670/DasanCallDial}{Huggingface}.

\bibliography{10_ref}

\input{9_appendix}

\end{document}

%% file: 0_abstract.tex
\begin{abstract}
Automatic Speech Recognition (ASR) technology is fundamental to customer service automation and large-scale transcription. However, even advanced ASR models exhibit inevitable errors in complex real-world environments such as call center conversations. When privacy restrictions preclude audio access, error correction must rely on text-based post-editing. Existing text-only approaches face significant challenges in low-resource languages, mainly due to a critical scarcity of annotated corpora and tailored correction methodologies. For Korean, this resource gap is particularly pronounced, as existing resources are predominantly designed for ASR training rather than text-based error correction. To address this, we introduce \textbf{DasanCallDial}, the first large-scale Korean benchmark dataset specifically curated for dialogue-level ASR error correction. Derived from genuine call center interactions, it comprises 1,974 dialogues with 115,460 utterances. Leveraging this resource, we propose \textbf{D}etector-Gated \textbf{C}ontextual \textbf{S}pan \textbf{C}orrection (\textbf{\textit{DCSC}}), a text-only post-editing framework for error-sparse Korean speech recognition transcripts. DCSC combines an encoder-based detector that first performs token-level error detection, followed by a language model-based corrector trained to rectify fine-grained span-level errors. Additionally, we employ dialogue-level context augmentation to enable the model to leverage discourse history for disambiguation. By employing multi-level granularity, our method achieves state-of-the-art performance, effectively overcoming the limitations of general LLMs in low-resource settings.
\end{abstract}

%% file: 1_introduction.tex
\section{Introduction}\label{sec:intro}
Automated Speech Recognition (ASR), often referred to as Speech-to-text (STT) technology, has become indispensable in domains such as customer support, accessibility services, and automated transcription. Recent advances in deep learning and natural language processing (NLP) have markedly boosted the accuracy of STT systems~\citep{10301513, AHLAWAT2025201}. However, in complex and dynamic environments such as call centers or consultant hotlines, speech recognizers still suffer from high error rates due to pronunciation variance, background noise, overlapping speech, and domain-specific jargon. Such misrecognitions disrupt effective complaint handling and public-service operations.

\begin{figure*}[t]
  \centering
  \includegraphics[width=\linewidth]{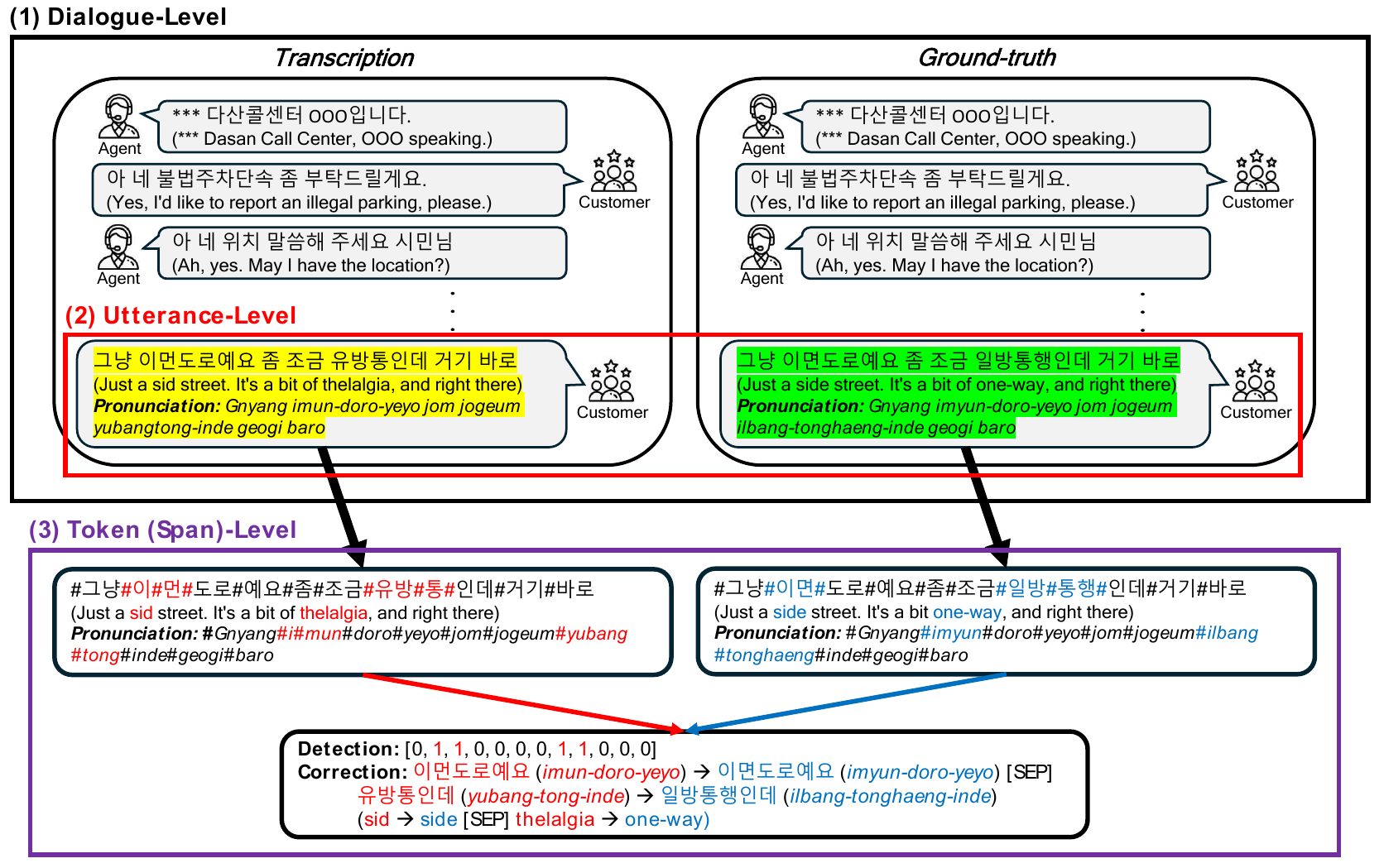}
  \caption{End-to-end illustration of the three-level granularity: \textbf{(1) Dialogue-level} shows a sample from the original dataset; \textbf{(2) Utterance-level} depicts utterances extracted from the dialogue; and \textbf{(3) Token(Span)-level} identifies token-level detection and span-level correction targets. Left and right examples represent noisy ASR transcriptions and human-annotated ground truth, respectively. In \textbf{(3)}, erroneous tokens are marked in \textcolor{MyRed}{red} and their corrections in \textcolor{MyBlue}{blue}.}
  \label{fig:intro}
\end{figure*}

Research on improving STT performance, therefore, spans two complementary fronts: (i) enhancing the acoustic and language components of the backbone ASR model~\citep{10.5555/3495724.3496768, gulati20_interspeech} and (ii) post-editing transcripts to fix residual errors. In public-service call logs, strict privacy regulations often prevent access to the raw audio, making text-only correction frameworks particularly valuable. Early work framed post-editing as monolingual ``translation'', fine-tuning sequence-to-sequence models to map noisy transcripts to clean text~\citep{9053051, DBLP:journals/corr/abs-2202-01157, ma23e_interspeech}. More recently, large language models (LLMs) have shown the ability to correct errors in context without task-specific tuning~\citep{ma2023generativelargelanguagemodels,10930744}. However, these successes are concentrated in high-resource languages such as English and Chinese; performance drops sharply in low-resource languages, where both error patterns and clean references are scarce~\citep{li24h_interspeech}.

To the best of our knowledge, the Korean ASR post-editing scenario—specifically involving full discourse-level analysis under the noisy, real-world conditions of call-center data—remains largely unexplored. To bridge this gap, we compile a new benchmark named \textbf{DasanCallDial}, comprising real call-center dialogues curated from the Seoul Dasan Call Foundation, an official government organization in Korea. The dialogues in {DasanCallDial} are automatically transcribed by a production ASR system and subsequently manually corrected to establish ground-truth references. \textbf{DasanCallDial} stands as the first dialogue-level Korean dataset for post-editing derived from raw, real-world environments. Uniquely, it features an error-sparse setting that accurately reflects the high baseline performance of modern STT models, distinguishing it from synthetic datasets with artificially inflated error rates. The final corpus contains approximately 1974 dialogue pairs with 115,460 utterance lines, with an error rate of 17.95\% across the distinct utterance set. 
%Detailed procedures and statistics are provided in Section~\ref{sec:dataset_construction}, and Section~\ref{sec:statistics} offers a comprehensive, multi-faceted analysis of the \textbf{DasanCallDial} dataset, including error type analyses. 
Leveraging this realistic benchmark, we propose \textbf{D}etector-Gated \textbf{C}ontextual \textbf{S}pan \textbf{C}orrection (\textbf{\textit{DCSC}}), a post-editing framework designed for error-sparse, text-only Korean ASR correction, and validate its effectiveness across multiple metrics.

Drawing on prior studies where introducing a dedicated detector before the corrector has been shown to improve correction accuracy~\citep{pu2024multistagelargelanguagemodel, yeen23_interspeech}, DCSC first applies a token-level detector to determine whether an utterance requires correction. If correction is needed, the full utterance is forwarded to a corrector rather than only the detected error tokens, allowing the corrector to revise the utterance using the surrounding lexical and syntactic context. Furthermore, inspired by findings that editing models perform best when trained on spans that truly require modification rather than on entire sentences~\citep{malmi-etal-2019-encode, stahlberg-kumar-2020-seq2edits, 10.1609/aaai.v37i11.26531}, we systematically \textbf{refine sample granularity in three levels}, as illustrated in Fig.~\ref{fig:intro}. At the beginning, the dataset comprises full \textbf{(1) dialogue-level} samples between call-center agents and customers. Then, each dialogue is split into individual \textbf{(2) utterance-level} samples, shortening context and increasing sample count (Section~\ref{sec:dial2utt}). We further refine each utterance by computing token-level edit distances between the ASR transcription and ground truth, yielding \textbf{(3) token-level} labels for detection (Section~\ref{sec:dectector}) and \textbf{(3) span-level} targets for correction (Section~\ref{sec:corrector}). Finally, to recover dialogue information lost during utterance-level segmentation, DCSC employs a \textbf{dialogue-level context augmentation} strategy (Section~\ref{sec:corrector}), concatenating the preceding discourse context with the target utterance.

We empirically validate the effectiveness of DCSC in Section~\ref{sec:main_results}. Specifically, the pkoT5-based DCSC achieves a detection accuracy (\textbf{Acc}) of 90.43, outperforming the zero-rule baseline of 83.43, and reduces the balanced word-error rate (\textbf{Bal-WER}) from 14.67 to 13.30. It also lowers the word-error rate on erroneous utterances (\textbf{E-WER}) from 29.35 to 26.27, demonstrating that DCSC performs meaningful corrections while keeping unnecessary edits on clean utterances low. Moreover, detailed ablation studies on the detector and corrector (Section~\ref{sec:detection_results}, Section~\ref{sec:correction_results}) show how token-level detection, span-level correction, dialogue context augmentation, and detector-guided routing each contribute to the final performance. Finally, we conduct extensive analyses to clarify why and when DCSC works effectively (Section~\ref{sec:analyses}). These include representation geometry, detector--corrector connectivity, external-domain transfer, speaker- and error-density-based breakdowns, and qualitative case studies. Together, these analyses characterize both the robustness and limitations of DCSC.

%% file: 2_related_work.tex
\section{Related work}

\subsection{Korean ASR datasets}

\begin{table*}[!htbp]
\scriptsize
\centering
\caption{Comparison of DasanCallDial and existing Korean ASR datasets based on content type, audio source environment, availability of ASR transcriptions, support for an error-sparse setting, and dataset scale.}
\resizebox{0.9\linewidth}{!}{
\begin{tabular}{l l l c c r}
\toprule
\textbf{Dataset}
& \textbf{Contents}
& \textbf{Audio Source}
& \textbf{STT Trans.}
& \textbf{Error-sparse}
& \textbf{\# Utterances} \\
\midrule

KsponSpeech~\citep{app10196936}
& Dialogue
& Controlled
& \xmark
& \xmark
& 625,545 \\

ClovaCall~\citep{ha20_interspeech}
& QA
& Controlled
& \xmark
& \xmark
& 61,000 \\

\midrule

KEBAP~\citep{koo-etal-2023-kebap}
& Sentence
& Controlled
& \cmark
& \xmark
& 2478 \\

Hyper-BTS~\citep{park-etal-2024-hyper}
& Sentence
& Synthetic
& \cmark
& \xmark
& 1,008,000 \\

\midrule

\textbf{DasanCallDial (Ours.)}
& \textbf{Dialogue}
& \textbf{Real World}
& \cmark
& \cmark
& 115,460 \\

\bottomrule
\end{tabular}
}
\label{tab:dataset_comparison}
\end{table*}

In the case of Korean, several large-scale speech datasets have been released. KsponSpeech~\citep{app10196936} provides 969 h of spontaneous speech and human-annotated transcriptions, recorded in a controlled environment to block external noise. Similarly, ClovaCall~\citep{ha20_interspeech} offers crowd-sourced recordings and source texts based on candidate QA pairs extracted from restaurant reservation scenarios. However, since these datasets are primarily designed for training ASR models, they do not provide the raw ASR output (STT transcript) texts. This omission renders them unsuitable for research on post-editing techniques where access to the raw audio source is restricted or where error correction must be performed solely within the text scope.

Consequently, there is a pressing need for datasets specifically designed for text-based ASR error correction. KEBAP~\citep{koo-etal-2023-kebap} constructed explainable data by analyzing and hypothesizing various error scenarios; however, its utility is limited by its sentence-level independence and the lack of realism due to the artificial post-insertion of external noise followed by transcription. Although Hyper-BTS~\citep{park-etal-2024-hyper} boasts a massive scale of 1M samples, it also consists of independent sentence-level data. Furthermore, it relies on synthetic generation pipelines—where existing text corpora (e.g., TED scripts or dictionaries) are processed through TTS models to generate speech, which is then converted back to text via STT—resulting in audio sources that are significantly synthetic. Moreover, existing Korean post-editing datasets consist exclusively of erroneous source samples and therefore do not provide the clean-majority label distribution needed to evaluate error-sparse post-editing.

In contrast, as shown in Table~\ref{tab:dataset_comparison}, our proposed \textbf{DasanCallDial} is the first Korean dataset to reflect real-world dialogue-level characteristics, collected by directly transcribing actual call center civil complaints. Furthermore, by retaining both correctly and incorrectly recognized raw ASR outputs without artificial filtering, it uniquely preserves the clean-majority, error-sparse distribution encountered in operational environments. We provide 115k samples of parallel data consisting of STT recognition outputs and corresponding human-annotated ground-truth texts, thereby facilitating the advancement of post-editing techniques specifically dedicated to correcting ASR errors.

\subsection{Post-editing for ASR transcriptions}
\label{sec:rel_post}
Automatic speech recognition (ASR) error detection is a long-standing post-processing step that flags recognition mistakes before they propagate to downstream applications. One common approach uses separate acoustic and transcript encoders. \citet{meripo22_interspeech} jointly encode acoustic and textual inputs, leveraging Wav2Vec~2.0~\citep{10.5555/3495724.3496768} for speech and BERT~\citep{devlin-etal-2019-bert} for text, and then train an entailment classifier to predict audio–transcript consistency. In many real-world settings, however, raw audio is not always accessible, which motivates text-only detection. \citet{10448230} therefore propose a lightweight utterance-level detector that estimates semantic distance to identify critical transcription errors without additional acoustic information. Moving beyond coarse utterance-level labels, \citet{gekhman-etal-2022-red} introduce a BERT-based sequence tagger that embeds ASR confidence scores alongside text to label each token as \textit{Error} or \textit{Non-Error}, yielding stronger detection performance.

Beyond detection, research on correcting ASR transcriptions has progressed along three intertwined axes: model capacity, pipeline structure, and supervision granularity. These dimensions collectively inform our approach. Early work shows that prompting GPT-style large language models (LLMs) can reduce word-error rate (WER) without fine-tuning~\citep{ma2023generativelargelanguagemodels, 10930744}, yet these gains are limited to high-resource languages. When parallel data are scarce, transformer sequence-to-sequence models fine-tuned on 1-best hypotheses provide a stronger baseline~\citep{9053051}. Subsequent studies add denoising objectives~\citep{DBLP:journals/corr/abs-2202-01157} or N-best conditioning~\citep{ma23e_interspeech}, demonstrating that modest corpora still yield competitive improvements when domain knowledge is injected explicitly. We therefore adopt Korean-pretrained seq2seq backbones as a realistic comparison against LLMs.

Beyond model capacity, the structural configuration of the pipeline and the granularity of supervision have emerged as critical factors in maximizing correction performance. Specifically, correction efficacy is significantly enhanced when error localization and rewriting are decoupled into distinct stages. These two-stage pipelines first tag erroneous spans and then edit only those regions, outperforming single-stage baselines under data scarcity~\citep{pu2024multistagelargelanguagemodel, yeen23_interspeech}. Concurrently, fine-grained supervision has proven beneficial: tag-and-realize editors~\citep{malmi-etal-2019-encode}, span-level transducers~\citep{stahlberg-kumar-2020-seq2edits}, and soft masking~\citep{10.1609/aaai.v37i11.26531} all show that models learn more precise edits when trained only on text that requires modification. Our dataset follows this progression, refining samples from dialogue to utterance and finally to token or span granularity, and applying them to an encoder-based detector and a seq2seq-based corrector.

\subsection{ASR post-editing in low-resource languages}
\label{sec:low_post_editing}
In contrast to the substantial advancements in ASR post-editing for high-resource languages such as English and Chinese, error correction research for low-resource languages has been constrained by data scarcity. Nevertheless, recent scholarship has actively pursued the construction of datasets and the development of post-editing methodologies specifically tailored to these resource-constrained environments.

For instance, \citet{anshor2025implementing} demonstrated the efficacy of rule-based approaches over LLMs for Sundanese. Their study reported a 25\% reduction in Word Error Rate (WER) by post-processing Whisper transcriptions using type-2 fuzzy logic rules. Similarly, for Rajasthani, \citet{10.1145/3793254} proposed a character-level alignment strategy that fuses outputs from heterogeneous recognizers (e.g., Whisper, MMS)—incorporating alignment gaps ($\emptyset$) as tokens—via a lightweight gated fusion mechanism. This method yielded lower error rates compared to both zero-shot LLMs (e.g., GPT-4o, Gemini 2.5 Pro) and small fine-tuned LLMs (e.g., Llama-3.2-3B). Regarding Latvian, \citet{znotins-gruzitis-2025-conversational} reported that a small-scale LLM with 2 billion parameters, when fine-tuned on a limited corpus, achieved sentence correction performance comparable to that of multi-billion parameter models.

Collectively, these findings suggest that in low-resource settings, problem-specific inductive bias and domain-adaptive training are more critical determinants of performance than mere model scaling. However, existing literature largely relies on naive training paradigms that directly map input sentences to ground-truth references. Consequently, the efficacy of advanced target designs—such as the utilization of finer granularity or two-stage pipelines—within low-resource contexts remains underexplored.

%% file: 3_dataset.tex
\section{Dataset}
\label{sec:dataset}

\subsection{Dataset construction}\label{sec:dataset_construction}

\subsubsection{Original dialogue-level dataset}\label{sec:dial_data}
To construct the dataset for ASR post-editing, we collect actual civil complaint call recordings from the Dasan Call Center, involving the participation of approximately 370 counselors. With an average call duration of 2 min and 30 s, the total volume of speech amounts to approximately 296,100 s (82 h and 15 min). The collected audio data are transcribed using \texttt{HAIV}, an AI-driven speech-to-text model optimized for cloud environments and operated on an on-premises infrastructure utilizing Docker and Kubernetes. Based on these transcriptions, we manually create corresponding corrected dialogues to serve as the ground truth by meticulously listening to the raw audio files. The annotation process is guided by extensive call center operational experience to ensure that the text is transcribed in the most contextually plausible manner. To assess whether the observed error-sparse characteristics are specific to \texttt{HAIV}, we additionally transcribe the same audio using a Korean-fine-tuned \texttt{Whisper} model and provide a cross-ASR comparison in \ref{app:cross_asr}.

Given that the dataset comprises real voice recordings between citizens and counselors, strict anonymization is essential to protect privacy. Therefore, we implement a de-identification process for both the ASR transcriptions and the ground-truth annotations. This process follows a structured protocol comprising three steps: (1) during the initial data extraction, all numerical values were automatically masked with asterisks (*) using SQL regular expressions; (2) the data from Step 1 were manually de-identified to remove personal names and other sensitive information; and (3) the processed data underwent a thorough final review to verify the completeness of the anonymization.

Two personnel with extensive long-term experience at the Dasan Call Center performed the annotation and anonymization tasks. One expert had 8 years of experience in constructing and operating AI-based consultation systems (e.g., STT and consultation assistants), while the other had 15 years of experience across a broader range of AI consultation systems, including ARS, STT, call-bots, and chat systems. The annotation workload was first partitioned between the two experts for independent labeling, followed by a rigorous cross-validation procedure to ensure data quality. During the first-pass annotation, 10,391 utterances are corrected, accounting for approximately 9\% of the full utterance set. The subsequent validation stage further refines these annotations: 914 first-pass corrections are reverted to the original transcription, and 1372 are revised into different corrected forms; 5253 utterances unchanged in the first pass are newly corrected. Importantly, these categories have different reference populations and therefore should not be aggregated relative only to the 10,391 initially corrected utterances. Among the initially corrected utterances, $(914 + 1{,}372)/10{,}391 \approx 22.0\%$ were either reverted or revised. In contrast, the 5253 newly identified errors correspond to approximately $5.0\%$ of the 105,069 utterances that had remained unchanged after the first pass. Overall, 7539 utterances, or approximately 6.53\% of the complete set of 115,460 utterances, underwent a change in either correction status or corrected form during the second-pass review. In instances of disagreement, final decisions are reached through consensus-based discussion. Spanning approximately eight months from raw audio collection to final dataset construction, this project ultimately processed 1974 call instances, yielding 115,460 utterances for both raw and corrected transcriptions. We provide detailed annotation guideline in \ref{app:anno_guide}.

To directly assess inter-annotator consistency, we additionally sampled 1000 utterances from the test set for independent double annotation. Each annotator independently assigned a binary label indicating whether an utterance contained an ASR error or was error-free. The resulting Cohen's $\kappa$ was 0.73, and among the utterances independently identified as erroneous by both annotators, the exact-match agreement between their corrected transcriptions was approximately 85.5\%. These results suggest consistent inter-annotator decisions in both error identification and corrected transcription generation.

\subsubsection{Dialogue-level to utterance-level}\label{sec:dial2utt}
We initially considered using the entire dialogue as input, hypothesizing that a broader conversational context would facilitate more accurate corrections. However, even large-scale language models often struggle to fully leverage lengthy contexts in low-resource languages such as Korean~\citep{aycock2025can}. Drawing on studies indicating that shorter contexts or multiple single-sentence examples can improve performance in low-resource scenarios~\citep{cahyawijaya-etal-2024-llms}, we refine the sample granularity from \textbf{dialogue-level} to \textbf{utterance-level} by splitting each dialogue into individual utterances.

To preserve the information in each original dialogue sample, we create a dataframe with [\textit{speaker, dialogue key, transcribed utterance, ground truth utterance}]. We define a \textit{dialogue key} by concatenating the recording date with a running index for that day, ensuring that all utterances originating from the same conversation share an identical key.

\subsubsection{DasanCallDial: Dataset specifications}
\label{sec:dasancalldial}

\begin{table*}[!htbp]
\centering
%\caption{Dataset statistics by split.}
\caption{Detailed statistics of the DasanCallDial benchmark organized by data split and granularity.}
\label{tab:dataset_stats}
\renewcommand{\arraystretch}{1.1}
\resizebox{0.8\linewidth}{!}{%
\begin{tabular}{l||rrr|rrr}
\toprule
\multirow{2}{*}{\textbf{Split}} & \multicolumn{3}{c|}{\textbf{Dialogue-level}} & \multicolumn{3}{c}{\textbf{Utterance-level}} \\
\cmidrule(lr){2-4}\cmidrule(lr){5-7}
& \textit{\# Samples} & \textit{Avg.\ Turns} & \textit{\# Utterances} & \textit{\# Samples} & \textit{Avg.\ Chars} & \textit{Error Rate (\%)} \\
\midrule
Total & 1974 & 58.49 & 115,460 & 64,710 & 22.44 & 17.95 \\
\midrule
Train & 1579 & 57.87 & 91,378  & 50,480 & 22.57 & 18.01 \\
Validation   & 98   & 49.11 &  4813   & 2963  & 21.14 & 13.26 \\
Test  & 297  & 64.88 & 19,269  & 11,267 & 22.20 & 16.57 \\
\bottomrule
\end{tabular}
}
\end{table*}

We propose our constructed \textbf{\textit{DasanCallDial}} dataset for both dialogue-level and utterance-level configurations. As detailed in Table~\ref{tab:dataset_stats}, the dialogue-level dataset comprises a total of 1974 call dialogues. Each dialogue instance consists of approximately 58.49 turns on average, resulting in a cumulative total of 115,460 utterance lines.
We first partition the dialogue instances into training, validation, and test sets with 0.80/0.05/0.15 ratio. Consequently, the training, validation, and test sets contain 1,579, 98, and 297 dialogue instances, respectively. These subsets correspond to 91,378, 4,813, and 19,269 utterances, with average turn counts of 57.87, 49.11, and 64.88 per dialogue.

However, raw call-center data frequently contain repeated phrases, such as common inquiries or filler words, which can skew model training. To address this, we constructed the utterance-level dataset by deduplicating overlapping utterances within each partition of the dialogue-level dataset. We carefully rebalanced duplicates within each subset to ensure that frequently used expressions remained adequately represented across the training, validation, and test sets.
This deduplication process yielded 50,480, 2,963, and 11,267 unique utterance instances for the respective subsets, with average character lengths of 22.57, 21.14, and 22.20. Notably, since a significant portion of the removed duplicates consisted of easily recognized, error-free common phrases, the resulting error rates for the filtered subsets increased to 18.01\%, 13.26\%, and 16.57\%. These figures are approximately double the original overall sample error rate of 9.83\%.
By removing redundant correct samples, this strategy mitigates the sparsity of erroneous samples in the training set. Furthermore, the resulting error distribution aligns with the operational accuracy of the deployed STT model. This approach ultimately offers a realistic reflection of everyday colloquial usage while preserving a balanced distribution of utterances for robust model training and evaluation.

\subsection{Dataset analysis}
\label{sec:statistics}

\subsubsection{Error type statistics}
\label{sec:type_statistics}

\begin{figure*}[!htbp]
\centering
\includegraphics[width=0.8\linewidth]{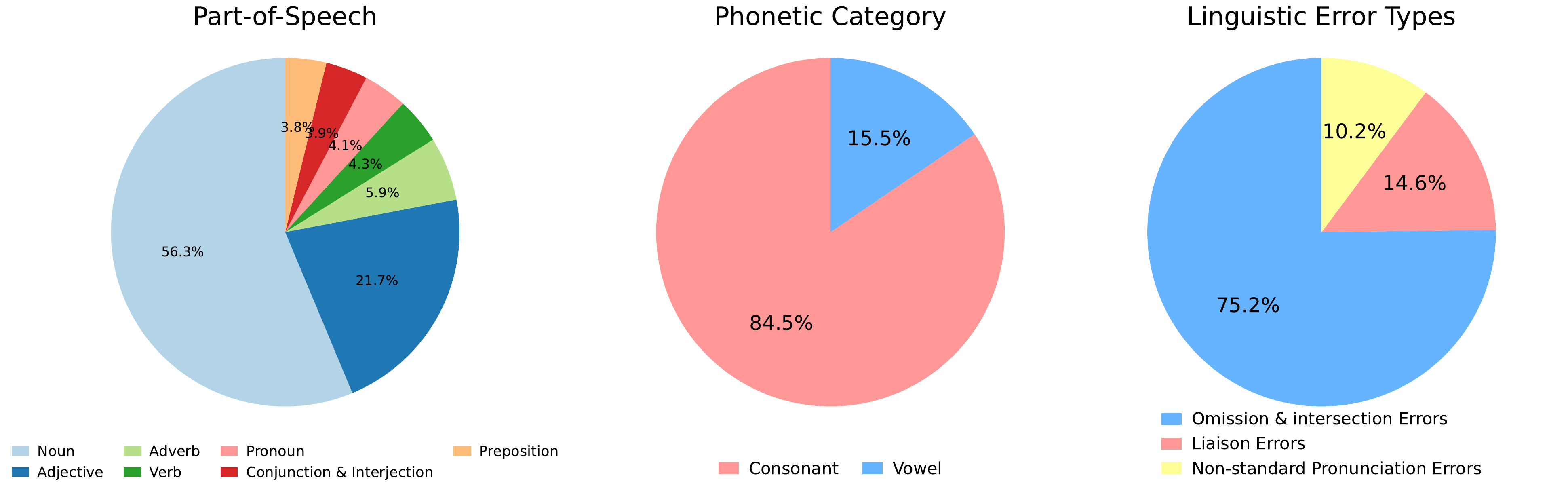} 
%\caption{Overall statistics of Seoul Dasan Call Foundation collected data.}
\caption{Distribution of ASR errors by part-of-speech (left), phonetic category (center) and linguistic error types (right) within the DasanCallDial dataset.}
\label{fig:error_type}
\end{figure*}

In terms of error attribution, a part-of-speech analysis in the left of Fig.~\ref{fig:error_type} shows that noun-related errors account for the largest proportion (56.3\%), followed by errors involving verbs (21.7\%), adjectives (5.9\%), prepositions (4.3\%), pronouns (4.1\%), adverbs (3.8\%), conjunctions (2.9\%), and interjections (1.0\%). For instance, noun errors frequently occur when words such as ``jeona'' are transcribed instead of ``jeonhwa (telephone)'' or ``charang'' instead of ``charyang (car)''. Verb errors arise when ``badumyeon'' is recognized in place of ``badeumyeon (if receive)'' or ``andogo'' instead of ``andoego (it does not work)''. Adjective errors appear when ``geureomeuro'' is transcribed instead of the correct form ``geureomuro (hence)'' or ``gareuchida'' is misrecognized as ``garikida (indicate)''.

From a phonetic perspective, a significant majority of errors (84.5\%) involve consonant-level distortions, whereas vowel-related errors account for the remaining 15.5\% as shown in the center of Fig.~\ref{fig:error_type}. For example, the word ``bab (rice)'' is often misrecognized as ``pap'', and ``seoul (Seoul)'' is transcribed as ``sseoul''. Concerning vowel confusions, ``yul'' may be interpreted as ``ul'', and ``e'' as ``ye''. Pronunciation-related errors can be further divided into three main categories, as shown in the right of Fig.~\ref{fig:error_type}. The most prevalent type is omission and insertion errors (75.2\%), in which phonemes are unintentionally dropped or added—for example, the phrase ``dongjakgu samuesil e'' is transcribed in place of the correct ``dongjakgucheong samusil (the Dongjak District Office)''. Liaison errors (14.6\%) stem from phonological links between adjacent syllables, as when ``seo ulsicheong'' is interpreted instead of ``seoulsicheong (Seoul City Hall)''. Finally, non-standard pronunciation errors (10.2\%) occur when non-canonical pronunciations are used, such as ``unsuhosa'' being recognized instead of ``unsuhoesa (transport company)''.

These statistics highlight the multifaceted nature of speech recognition errors in Korean, spanning lexical, phonological, and pronunciation-level variations. Accordingly, sophisticated error-correction mechanisms tailored to Korean linguistic characteristics are required. A corresponding analysis based on \texttt{Whisper} transcriptions is provided in \ref{app:cross_asr}, where we compare the error rate and detailed error distributions across the two ASR systems.

\subsubsection{Speaker type analysis}
\label{sec:speaker_analysis}

\begin{table}[!h]
\centering
\caption{Dataset statistics across two speaker types: Counselors \& Customers.}
\label{tab:speaker_anal}
\renewcommand{\arraystretch}{1.1}
\resizebox{\columnwidth}{!}{%
\begin{tabular}{l||rr|rr}
\toprule
\multirow{2}{*}{\textbf{Split}} & \multicolumn{2}{c|}{\textbf{Counselors}} & \multicolumn{2}{c}{\textbf{Customers}} \\
\cmidrule(lr){2-3}\cmidrule(lr){4-5}
& \textit{\# Samples} & \textit{Error Rate (\%)} & \textit{\# Samples} & \textit{Error Rate (\%)} \\
\midrule
Total & 33,542 & 13.47 & 31,168 & 22.71 \\
\midrule
Train & 26,179 & 13.56 & 24,301 & 22.80 \\
Validation   & 1547 & 10.28 &  1416   & 16.53  \\
Test  & 5816  & 12.41 & 5451  & 21.01 \\
\bottomrule
\end{tabular}
}
\end{table}

Fundamentally, the \textbf{DasanCallDial} dataset comprises dyadic conversations between a counselor and a customer. In this section, we present a statistical analysis of utterance counts and error rates by speaker role within the deduplicated utterance-level configuration. As shown in Table~\ref{tab:speaker_anal}, the dataset contains a total of 33,542 counselor utterances, exhibiting an error rate of 13.47\%. Conversely, customer utterances total 31,168—slightly fewer than those of the counselors—but display a significantly higher error rate of 22.71\%. 
%Across data partitions, both the training and test sets maintain distributions consistent with these overall figures.

We infer that this disparity arises because counselor speech is relatively standardized and articulated with professional clarity, facilitating accurate ASR transcription. In contrast, customer utterances are unstructured, varying widely depending on the specific complaint, and often feature less precise articulation, resulting in lower recognition accuracy. These statistical properties appropriately reflect the realistic performance characteristics of ASR systems operating in actual call center environments.

\subsubsection{Topic analysis}
\label{sec:topic}

\begin{figure}[!htbp]
\centering
\includegraphics[width=\columnwidth]{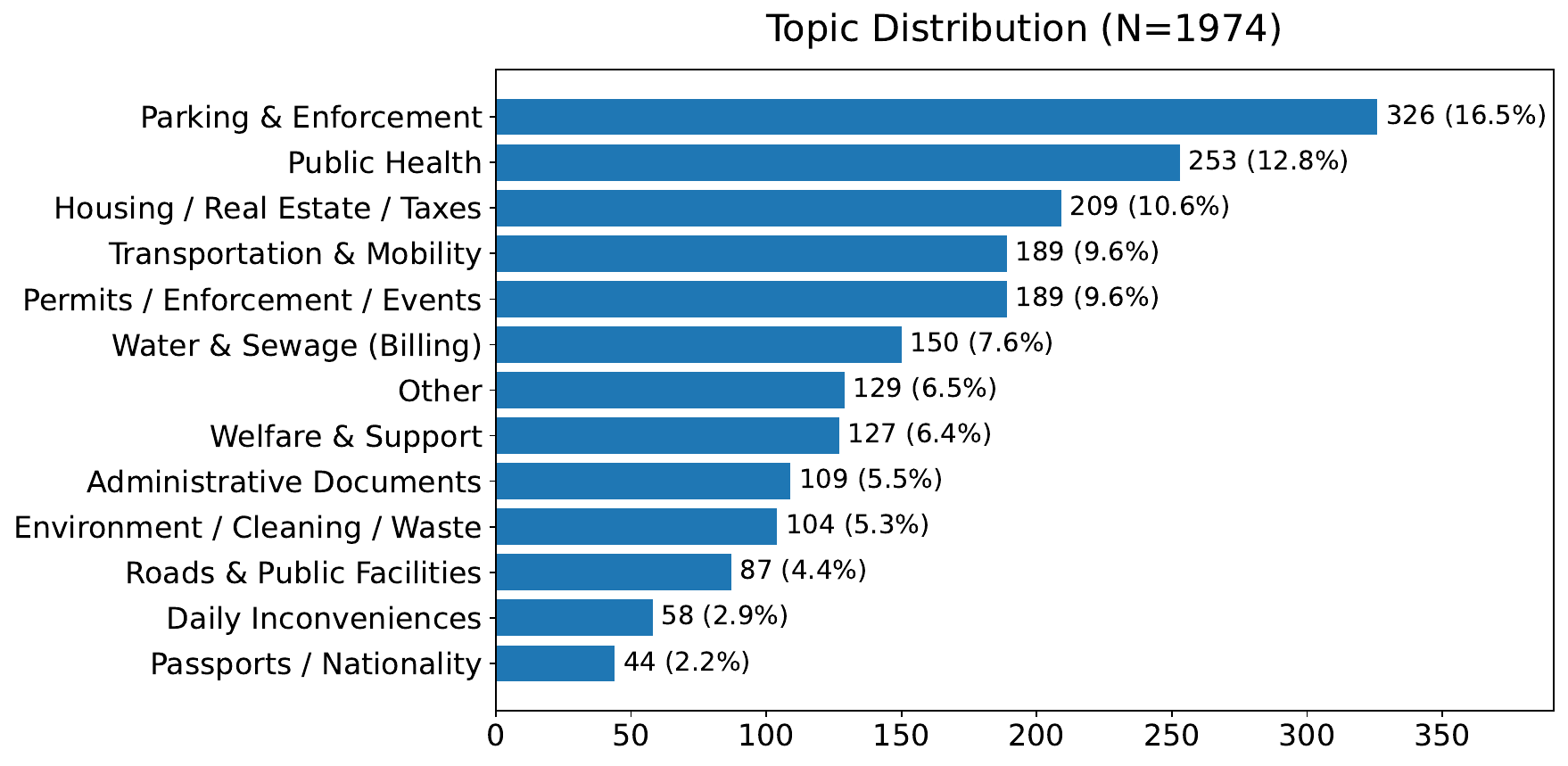} 
\caption{Overall Topic Distribution of \textbf{DasanCallDial}.}
\label{fig:topic}
\end{figure}

As a comprehensive welfare agency serving the public, the Dasan Call Center receives a diverse spectrum of civil complaints. To effectively handle these varied inquiries, the organization is divided into specialized departments. Our \textbf{DasanCallDial} dataset is curated with a focus on 12 primary departments: \textit{Parking \& Enforcement, Transportation \& Mobility, Public Health (Clinic), Permits / Enforcement / Events, Welfare \& Support, Roads \& Public Facilities, Water \& Sewage (Billing), Housing / Real Estate / Taxes, Administrative Documents, Daily Inconveniences, Environment / Cleaning / Waste, and Passports / Nationality}. We examine the topic distribution by aligning each dialogue instance with its corresponding department. As illustrated in Fig.~\ref{fig:topic}, the dataset exhibits a relatively balanced distribution, with no single department exceeding 20\% of the total volume. The \textit{Parking \& Enforcement} department accounts for the largest share, contributing 326 instances (16.5\%). In contrast, the \textit{Passports / Nationality} department represents the smallest share, with 44 instances (2.2\%). Data originating from departments other than the aforementioned 12 are categorized as ``Other'', totaling 129 instances (6.5\%). This broad topical coverage indicates that our dataset possesses high generalizability, facilitating its flexible application across multiple domains.

\subsubsection{Frequency analysis}
\label{sec:frequency}

\begin{figure*}[!htbp]
\centering
\includegraphics[width=0.8\linewidth]{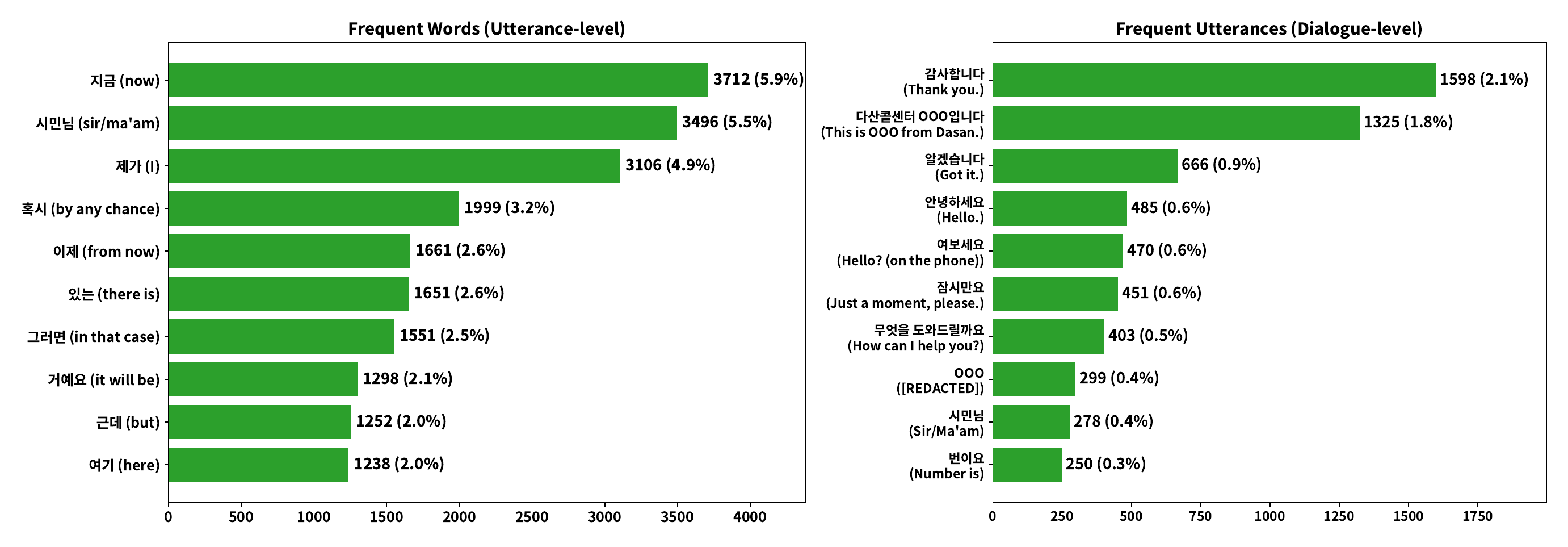} 
\caption{Top 10 Frequent Words / Utterances in \textbf{DasanCallDial}.}
\label{fig:frequency}
\end{figure*}

As noted in Section~\ref{sec:dasancalldial}, real-world consultation dialogues are characterized by a high recurrence of standardized expressions and a prevalence of colloquialisms, including filler words and sentence fragments. Accordingly, analyzing the frequency distribution of words and utterances provides a critical foundation for optimizing future post-processing tasks. We identified the top 10 most frequent words within the utterance-level dataset and the top 10 most frequent utterances within the dialogue-level dataset. As illustrated in Fig.~\ref{fig:frequency}, the word-level analysis reveals a predominance of temporal adverbs (e.g., ``now'', ``from now''), honorific titles (e.g., ``citizen''), and discourse markers (e.g., ``by any chance'', ``but'') intended to soften the conversational tone. In the case of frequent utterances, the distribution is dominated by standardized service scripts, including greetings, self-introductions, and closings (e.g., ``Thank you'', ``This is OOO from Dasan Call Center''). Furthermore, the presence of short, functional phrases (e.g., ``Just a moment'') and fragmented segments (e.g., ``Number is..''.) authentically reflects the spontaneous and interactive nature of spoken communication in a call center environment. However, notwithstanding the inevitable recurrence of such common Korean expressions, the cumulative frequency of the top 10 items accounts for merely 33.3\% and 8.2\% of the respective totals. This limited coverage suggests that the dataset successfully secures a wide diversity of expressions beyond these high-frequency patterns.

%% file: 4_methodology.tex
\section{Methodology}
%\section{Two-Stage Error Correction Framework}
\label{sec:method}
In this section, we present \textbf{D}etector-Gated \textbf{C}ontextual \textbf{S}pan \textbf{C}orrection (\textbf{\textit{DCSC}}), a post-editing framework that leverages the \textbf{DasanCallDial} dataset described in Section~\ref{sec:dataset}. DCSC operates as a two-stage pipeline: first, an encoder-based detector localizes recognition errors within raw transcriptions and determines whether correction is necessary; subsequently, a Seq2Seq-based or LLM-based corrector receives the full error-flagged utterance with dialogue context and generates span-level correction instructions. The primary objective of DCSC is to guide the model to produce edited transcriptions that align precisely with the annotated ground truth while minimizing unnecessary modifications to already-correct utterances.

\subsection{ASR error detection}
\label{sec:dectector}

\subsubsection{Data preprocessing: Utterance-level to token-level}\label{sec:utt2tok}
Let $U$ denote an \textbf{utterance-level} transcription instance and $G$ its \textbf{utterance-level} ground-truth. Rather than simply determining whether $U$ contains any errors, we refine the task to detect errors in each token of $U$, as in sequence-masking tasks. Specifically, we utilize a tokenizer of encoder-based language model to convert $U$ and $G$ into \textbf{token-level} sequences $U_{dec} = \{\hat{t}_1, \ldots, \hat{t}_n\}$ and $G_{dec} = \{t_1, \ldots, t_{n'}\}$, where $n$ and ${n'}$ are their respective lengths. We then compute the edit distance between these tokenized representations using the \texttt{SequenceMatcher} from the \texttt{difflib} library. Any token in $U_{dec}$ that differs from $G_{dec}$ (because of replacement, deletion, or insertion) is masked with 1, while matched tokens are masked with 0. If an insertion occurs, the token immediately preceding the insertion point is masked; if the insertion is at the start of the utterance, the token immediately following that point is masked. As a result, we obtain a \textbf{token-level} mask sequence $M = \{m_1, \ldots, m_n\}$, where $m_i = 1$ when $\hat{t}_i$ is erroneous and $0$ otherwise.

\subsubsection{Training detector}\label{sec:train_dec}
We formulate the detection task as a sequence-labeling problem~\citep{gekhman-etal-2022-red}, where each token in $U_{dec}$ is assigned a binary mask. To develop a model for this task, we fine-tune the same model used in the tokenization phase (Section~\ref{sec:utt2tok}) so that it outputs the mask sequence $M$ given $U_{dec}$. The loss function is defined as:
\begin{multline*}
\mathcal{L} = -\frac{1}{N}\sum_{i=1}^{N} \frac{1}{n^{(i)}}\sum_{t=1}^{n^{(i)}} \Bigl( \\
\lambda\, m_t^{(i)} \log\bigl(p(m_t^{(i)})\bigr) \\
+ \bigl(1-m_t^{(i)}\bigr)\log\bigl(1-p(m_t^{(i)})\bigr) \Bigr)
\end{multline*}
where $N$ is the number of training samples. For the $i$-th input $U_{dec}^{(i)}$, $n^{(i)}$ denotes its token length, $m_t^{(i)}$ the binary mask for the $t$-th token, and $p(m_t^{(i)})$ the predicted probability of assigning an error mask (1) to that token. Because non-erroneous tokens greatly outnumber erroneous ones (Section~\ref{sec:statistics}), we introduce a weight factor $\lambda = 8$ to emphasize learning on the sparse error tokens.

\subsection{ASR error correction}\label{sec:corrector}

\subsubsection{Data preprocessing: Utterance-level to span-level}\label{sec:utt2span}
To correct errors within $U$, we assume that the model learns more effectively when it focuses not on the entire sentence but specifically on the segments that require repair and their corrected forms. A complete sentence may admit many potential edits; narrowing attention to the precise spans to be modified simplifies the editing objective~\citep{stahlberg-kumar-2020-seq2edits, 10.1609/aaai.v37i11.26531}. We therefore refine the \textbf{utterance-level} ground truth $G$ into a \textbf{span-level} representation that explicitly indicates which segments need modification and how they should be corrected.

Following the notation in Section~\ref{sec:utt2tok}, let $U_{cor} = \{\hat{t}_1, \hat{t}_2, \ldots, \hat{t}_n\}$ be the tokenized form of $U$, and let $G_{cor} = \{t_1, t_2, \ldots, t_m\}$ be the tokenized form of $G$. Using the same sequence-masking strategy described in Section~\ref{sec:train_dec}, we measure the edit distance between $U_{cor}$ and $G_{cor}$ to determine which tokens in $U_{cor}$ must be replaced, inserted, or deleted. These tokens constitute the error spans $\textit{span}^U$, while the corresponding correct tokens constitute $\textit{span}^G$. In contrast to the detection phase, we treat tokens at the word level and merge consecutive tokens into a single span when warranted—for example, for insertions, deletions, or contiguous erroneous segments.

We then pair each span in $\textit{span}^U$ with its counterpart in $\textit{span}^G$ using an arrow token ($\rightarrow$) and concatenate multiple pairs with the learnable $[\text{SEP}]$ token, creating the final \textbf{span-level} label $\mathcal{S}$. If no errors are present in $U$, we simply designate the phrase \texttt{``No Error''} in Korean as the target. Formally:
{\small
\begin{equation*}
\mathcal{S} =
\begin{cases}
\texttt{No Error}, & \text{if } s = 0, \\[4pt]
\begin{aligned}[t]
  &\text{span}_1^U \rightarrow \text{span}_1^G \,[\text{SEP}]\, \\
  &\text{span}_2^U \rightarrow \text{span}_2^G \,[\text{SEP}]\, \cdots,
\end{aligned} & \text{otherwise}
\end{cases}
\end{equation*}}
Here, $s$ is the number of extracted spans, while $\text{span}_i^{U}$ and $\text{span}_i^{G}$ denote the $i$-th erroneous span in $\textit{span}^{U}$ and its corrected form in $\textit{span}^{G}$, respectively.

\subsubsection{Adopting dialogue-level context augmentation}
\label{sec:con_aug}
While operating at the utterance level offers significant potential by reducing the model's input length and concretizing the scope of correction, it inherently suffers from a critical limitation: the inability to resolve errors that necessitate dialogue context for disambiguation. To address this, we employ a dialogue-level context augmentation strategy, concatenating prior discourse information with the target utterance $U$.

Let a transcribed dialogue instance be denoted as $D_U = \{U_1, U_2, \ldots, U_N\}$ and the corresponding ground-truth dialogue instance as $D_G = \{G_1, G_2, \ldots, G_N\}$, where $N$ represents the total number of turns in the dialogue. When processing a target utterance $U_i$ for editing, we retrieve the up to 10 immediately preceding ground-truth utterances from $D_G$ to serve as context.

To effectively distinguish between the reference dialogue context and the target utterance to be corrected, we introduce trainable special separator tokens, \texttt{[Dialogue Context]} and \texttt{[Target Utterance]}. These are embedded into the sequence to construct the final input $U_{\text{context}}$. The formulation of $U_{\text{context}}$ is expressed as follows:

\begin{equation*}
\begin{split}
U_{\text{context}} = (\;
  & \texttt{[Dialogue Context]} \,;\, \\
  & G_{\max(1, i-10):i-1} \,;\, \\
  & \texttt{[Target Utterance]} \,;\, U_i \;)
\end{split}
\end{equation*}

\subsubsection{Training corrector}\label{sec:train_cor}
We fine-tune a corrector model to predict the \textbf{span-level} representation $\mathcal{S}$ given an utterance-level transcription $U$. At inference time, the predicted $\hat{\mathcal{S}}$ is not a complete sentence, so we apply a post-processing step to construct the corrected utterance $\hat{G}$. Specifically, we split $\hat{\mathcal{S}}$ at each $[\text{SEP}]$ token and replace every $\hat{\text{span}}_i^{U}$ in $U$ with its corresponding $\hat{\text{span}}_i^{G}$. If the model outputs \texttt{No Error}, we simply return $U$ unchanged.

Under the adopting context augmentation, $U_{\text{context}}$ serves as the input during the training phase, replacing the isolated utterance $U$. The provided dialogue context is utilized exclusively for reference purposes; the model is optimized to generate only $\mathcal{S}$ of the target utterance.

\subsection{Combining detector and corrector}
\label{sec:dec_cor}

\begin{figure*}[!htbp]
  \centering
  \includegraphics[width=0.8\linewidth]{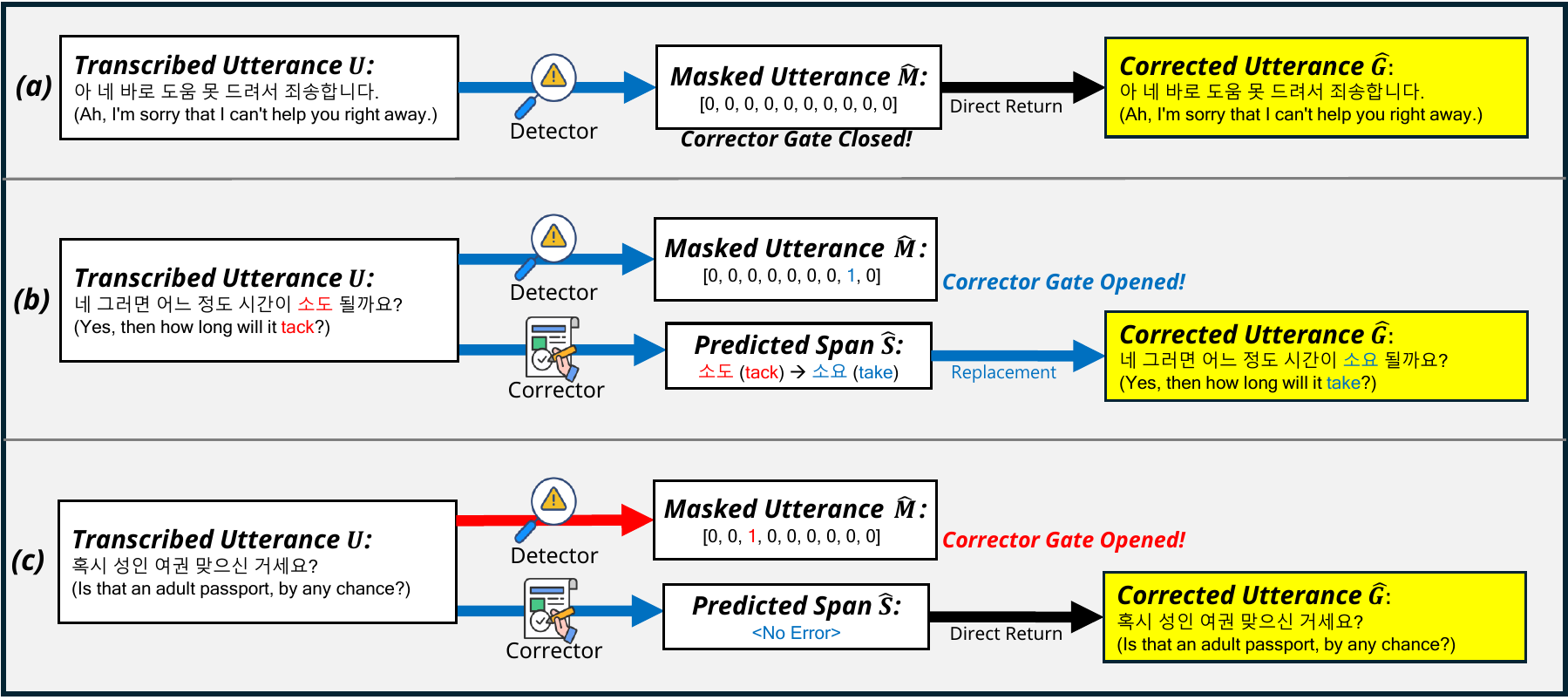}
  \caption{Illustration of the DCSC inference pipeline. Blue arrows denote correct actions; red arrows indicate mistakes. (a) Detector sees no error, the gate stays closed, and $U$ is returned. (b) Detector opens the gate and the corrector repairs the utterance. (c) Detector raises a false positive, but the corrector detects no error and echoes $U$.}
  \label{fig:enc+seq}
\end{figure*}

The final component of \textbf{\textit{DCSC}} combines the detector and corrector into a gated inference pipeline. Inspired by prior work demonstrating that a detector-corrector pipeline can improve ASR error correction~\citep{pu2024multistagelargelanguagemodel, yeen23_interspeech}, we integrate the detector from Section~\ref{sec:dectector} into our seq2seq approach.

First, for each utterance-level training sample $U_{\text{train}}$, we obtain a predicted mask sequence $\hat{M}{\text{train}}$ from the detector: if $\hat{M}{\text{train}}$ contains at least one 1, $U_{\text{train}}$ is flagged as erroneous; otherwise, it is flagged as error-free. We then assemble a new training set for the seq2seq corrector by collecting
\begin{enumerate}
\item All samples whose $M_{\text{train}}$ actually contain at least one erroneous token (1),
\item All samples that the detector incorrectly flags as erroneous, even though $U_{\text{train}}$ is error-free, and
\item A random selection of error-free utterances, balanced to match the quantity of the erroneous sample group.
\end{enumerate}
The corrector model is further fine-tuned on this curated set following the fine-grained scheme in Section~\ref{sec:train_cor}. This refinement enables the model to focus on genuine correction behavior while preventing spurious detections from propagating downstream.

At inference time, we first apply the detector to an utterance-level transcription $U$. As illustrated in Fig.~\ref{fig:enc+seq}(a), when the detector deems $U$ error-free, the gate to the corrector remains closed and $U$ is returned unchanged. Otherwise, the detector flags an error and opens the gate to the corrector, forwarding $U$ to be refined into $\hat{G}$. Here, the detector's token-level predictions are used only for gating: the corrector receives the full utterance, rather than only the predicted error tokens, and independently determines the necessary revision. Fig.~\ref{fig:enc+seq}(b) shows a typical success case in which the detector identifies the error span and the corrector produces an accurate revision. Even when the detector mistakenly opens the gate for an error-free sentence, as in Fig.~\ref{fig:enc+seq}(c), the corrector can recognize that no change is required and simply echoes $U$, thereby compensating for the detector's false positive. Crucially, the dialogue-level context augmentation is applied exclusively to the corrector phase following detector inference. When the detector flags an utterance $U$ and forwards it for correction, we construct the context-augmented input $U_{\text{context}}$ instead of using the raw $U$. Since access to the ground-truth history $D_G$ is unavailable during inference, the context slot $G_{\max(1, i-10):i-1}$ within $U_{\text{context}}$ is populated autoregressively using the previously corrected utterances $\hat{G}_{\max(1, i-10):i-1}$.

%% file: 5_experiments.tex
\section{Experiments}
\label{sec:results_analysis}

%\subsection{Main Results}
\subsection{Comparative analysis of proposed post-editing method}
\label{sec:main_results}

\subsubsection{Setup}
\label{sec:main_setup}

\paragraph{Evaluation metrics for detection}
We assess error detection performance by evaluating the model's ability to correctly determine whether an utterance-level transcription $U$ requires correction. We formulate this as a binary classification task: a predicted label of 0 (error-free) is assigned if the system returns $U$ unchanged, and 1 (erroneous) otherwise. Correspondingly, the ground-truth is defined as 0 when $U$ exactly matches the ground-truth reference $G$, and 1 otherwise. Comparing these two labels yields \textbf{Precision, Recall, F1} and \textbf{Accuracy (Acc.)} on the test set. Given the pronounced class imbalance in Section~\ref{sec:dasancalldial}, we also report \textbf{Balanced Accuracy (Bal-Acc.)} for a more equitable measure.

\paragraph{Evaluation metrics for correction}
We evaluate how effectively the corrector transforms transcription $U$ into its revised form $\hat{G}$ with respect to the ground truth $G$. To gauge the model's proficiency in fixing errors, we track:
\begin{itemize}
\item \textbf{Exact Match Ratio (EM)}: The proportion of samples for which $\hat{G}$ exactly matches $G$.
\item \textbf{Balanced Exact Match Ratio (Bal-EM)}: \textbf{EM} is computed separately for genuinely error-free and erroneous utterances, and the two results are averaged.
\item \textbf{Balanced Word-Error-Rate (Bal-WER)}: We compute the Levenshtein distance~\citep{6ff6bbca842f4a909c6ea0d3cce748da} between $\hat{G}$ and $G$, normalizing by the number of words in $G$, yielding the usual Word-Error-Rate (WER). Because most tokens in transcriptions are already correct, averaging WER across all samples yields a near-zero value, making comparisons among baselines difficult. Consequently, we compute WER separately for normal (\textbf{N-WER}) and erroneous (\textbf{E-WER}) subsets and average these values to obtain \textbf{Bal-WER}. This metric offers a softer, more nuanced assessment than strict \textbf{Bal-EM} evaluation.
\end{itemize}

\paragraph{Backbone models}
For the backbone detector in our experiments, we utilize \texttt{koelectra-base-v3-discriminator} (\texttt{KoElectra})~\citep{park2020koelectra}, a model pre-trained on a large-scale Korean corpus using the ELECTRA architecture. ELECTRA employs a Replaced Token Detection (RTD) objective~\citep{Clark2020ELECTRA}, where the discriminator learns to distinguish between ``real'' tokens from the original text and ``fake'' tokens generated by a generator. This mechanism closely aligns with our token-wise error detection task, suggesting that the model can effectively transfer its learned representations to identify erroneous tokens with high accuracy. For the corrector models, we employ a diverse range of architectures, spanning from Korean-pretrained Sequence-to-Sequence (K-Seq2Seq) models to widely-used LLMs. As K-Seq2Seq baselines, we utilize \texttt{google/mt5-base} (\texttt{mT5})~\citep{xue-etal-2021-mt5}, \texttt{paust/pko-t5-base} (\texttt{pkoT5})~\citep{paust_pkot5_v1}, and \texttt{hyunwoongko/kobart} (\texttt{KoBart}). For LLMs, we adopt \texttt{Llama-3.1-8b-Instruct} (\texttt{Llama-3.1})~\citep{grattafiori2024llama3herdmodels} and \texttt{Qwen2.5-7b-Instruct} (\texttt{Qwen2.5})~\citep{qwen2025qwen25technicalreport}.

\paragraph{Baselines}
We compare our DCSC framework against several representative baselines commonly adopted in low-resource ASR post-editing, covering rule-based, character-level, and utterance-level post-editing strategies.

\begin{itemize}
\item \textit{Zero-rule}: This baseline directly accepts the original ASR transcription as the final output without applying any correction. It serves as a lower-bound reference for assessing whether a post-editing model provides a meaningful improvement over the deployed ASR system. Be aware that this baseline cannot yield \textbf{Precision}, \textbf{Recall}, and \textbf{F1} in the detection task.

\item \textit{IT2-FLS}~\citep{anshor2025implementing}: We adapt the interval Type-2 fuzzy logic system (IT2-FLS) as a backbone-free rule-based baseline. The method compares ASR tokens with lexicon candidates using phonetic similarity and applies fuzzy inference to decide whether to replace them. For Korean adaptation, we compute phonetic similarity with weighted edit distance over decomposed Hangul jamo. Following the original work, we use Gaussian interval Type-2 membership functions with Mamdani inference and Karnik--Mendel type reduction via \texttt{PyIT2FLS}~\citep{haghrah2025pyit2fls}, and tune the membership parameters and decision thresholds using a genetic algorithm with Bal-WER as the fitness objective.

\item \textit{Char-Seq2Seq}~\citep{10.1145/3793254}: We implement a character-level attentional encoder--decoder baseline. Although the original work proposes cross-view gated fusion using multiple ASR hypotheses from different recognizers, our dataset contains only a single ASR output for each utterance. We therefore adopt its single-view character-level sequence-to-sequence baseline and adapt it to Korean by treating Hangul syllables as character units. Following the original work, we train two-layer bidirectional LSTM encoder and a two-layer LSTM decoder with Bahdanau attention. It receives the ASR syllable sequence as input and directly generates the corrected syllable sequence.

\item \textit{Uttr2Uttr}~\citep{znotins-gruzitis-2025-conversational}: This baseline represents the standard utterance-level sequence-to-sequence post-editing paradigm. Unlike our proposed framework, it does not use a detector, dialogue context, or intermediate change-string representation. Instead, it directly maps a single ASR utterance to the corresponding full ground-truth sentence. We employ two backbone models: Llama-3.1 and pkoT5.
\end{itemize}

\paragraph{Implementation details}
When training detectors, we set the learning rate to \texttt{2e-5}, while correctors are trained with a learning rate of \texttt{1e-4}. Both model categories utilize a batch size of 24 and are optimized using the AdamW optimizer~\citep{loshchilov2018decoupled}. Training is conducted for 6 epochs for LLMs and 15 epochs for other models. We select the best-performing checkpoint based on the best validation scores: \textbf{F1} for detectors and \textbf{Bal-WER} for correctors. For corrector models, we apply a weight decay of \texttt{1e-3} during training and utilize greedy decoding during inference. For LLM fine-tuning, we employ LoRA~\citep{hu2022lora} and configure DeepSpeed ZeRO-3~\citep{10.1145/3394486.3406703} with a gradient accumulation step of 8. Regarding computational time, the average duration required to train and save the best checkpoint was approximately 1 h and 30 min for detectors, 2 h and 20 min for K-Seq2Seq correctors, and 7 h and 20 min for LLM correctors. All experiments were conducted on a cluster equipped with three NVIDIA RTX 6000 Ada GPUs, with each task allocated 30 GB of system memory and 4 CPU cores. Further configuration details are provided in \ref{appendix:exp_setup}.

\subsubsection{Results}

\begin{table*}[t]
\centering
\caption{End-to-end performance of DCSC and baseline post-editing methods across all corrector backbones. DCSC variants are highlighted with gray rows. Metrics with $(\uparrow)$ indicate higher is better, while those with $(\downarrow)$ indicate lower is better. Best results are in \textbf{bold}. Changes relative to the zero-rule baseline are shown in parentheses, where red indicates performance improvement and blue indicates degradation for each metric. Models marked with $^{\dagger}$ are our primary configurations and are reported as averages over three random seeds.}
\label{tab:main_results}
\renewcommand{\arraystretch}{1.2}
\setlength{\tabcolsep}{4pt}
\resizebox{\linewidth}{!}{%
\begin{tabular}{l|ccccc|ccccc}
\toprule
\multirow{2}{*}{\textbf{Model}} 
& \multicolumn{5}{c|}{\textbf{Detection}} 
& \multicolumn{5}{c}{\textbf{Correction}} \\
\cmidrule(lr){2-6} \cmidrule(lr){7-11}
  & \textbf{Precision} \(\uparrow\) & \textbf{Recall} \(\uparrow\) & \textbf{F1} \(\uparrow\) & \textbf{Acc} \(\uparrow\) & \textbf{Bal-Acc} \(\uparrow\)
  & \textbf{EM} \(\uparrow\) & \textbf{Bal-EM} \(\uparrow\) & \textbf{N-WER} \(\downarrow\) & \textbf{E-WER} \(\downarrow\) & \textbf{Bal-WER} \(\downarrow\) \\
\midrule

\textit{Zero-rule}
   & -- & -- & -- & 83.43 & 50.00
   & 83.43  & 50.00 & 0 & 29.35 & 14.67 \\
\midrule
\midrule
\textit{IT2-FLS}
   & 28.57 & 2.06 & 3.84 & 82.92 {\footnotesize \textcolor{blue}{(-0.51)}} & 50.52 {\footnotesize \textcolor{red}{(+0.52)}}
   & 82.74 {\footnotesize \textcolor{blue}{(-0.69)}} & 49.84 {\footnotesize \textcolor{blue}{(-0.16)}} & 0.09 {\footnotesize \textcolor{blue}{(+0.09)}} & 28.70 {\footnotesize \textcolor{red}{(-0.65)}} & 14.40 {\footnotesize \textcolor{red}{(-0.27)}} \\
   
\textit{Char-Seq2Seq}
   & 43.32 & 7.26 & 12.44 & 83.06 {\footnotesize \textcolor{blue}{(-0.37)}} & 52.69 {\footnotesize \textcolor{red}{(+2.69)}}
   & 82.29 {\footnotesize \textcolor{blue}{(-1.14)}} & 49.52 {\footnotesize \textcolor{blue}{(-0.48)}} & 0.19 {\footnotesize \textcolor{blue}{(+0.19)}} & 29.08 {\footnotesize \textcolor{red}{(-1.27)}} & 14.64 {\footnotesize \textcolor{red}{(-0.03)}} \\

\textit{Llama-3.1 + Uttr2Uttr}
   & 11.63 & 46.88 & 18.64 & 32.17 {\footnotesize \textcolor{blue}{(-51.26)}} & 38.06 {\footnotesize \textcolor{blue}{(-11.94)}}
   & 51.34 {\footnotesize \textcolor{blue}{(-32.09)}} & 33.79 {\footnotesize \textcolor{blue}{(-16.21)}} & 83.29 {\footnotesize \textcolor{blue}{(+83.29)}} & 44.09 {\footnotesize \textcolor{blue}{(+14.74)}} & 63.69 {\footnotesize \textcolor{blue}{(+49.02)}} \\

\textit{pkoT5 + Uttr2Uttr}
   & 55.83 & 20.96 & 30.47 & 84.15 {\footnotesize \textcolor{red}{(+0.72)}} & 58.83 {\footnotesize \textcolor{red}{(+8.83)}}
   & 65.10 {\footnotesize \textcolor{blue}{(-18.33)}} & 42.34 {\footnotesize \textcolor{blue}{(-7.66)}} & 22.17 {\footnotesize \textcolor{blue}{(+22.17)}} & 28.49 {\footnotesize \textcolor{red}{(-0.86)}} & 25.33 {\footnotesize \textcolor{blue}{(+10.66)}} \\
   
\midrule
\midrule
\rowcolor{gray!20}
\textit{Llama-3.1 + DCSC (ours.)$^{\dagger}$}
   & 74.81
   & 69.92
   & 72.28
   & 90.07 {\footnotesize \textcolor{red}{(+6.64)}}
   & 80.73 {\footnotesize \textcolor{red}{(+30.73)}}
   & 85.05 {\footnotesize \textcolor{red}{(+1.62)}}
   & \textbf{53.89 {\footnotesize \textcolor{red}{(+3.89)}}}
   & 0.25 {\footnotesize \textcolor{blue}{(+0.25)}}
   & 26.56 {\footnotesize \textcolor{red}{(-2.79)}}
   & 13.40 {\footnotesize \textcolor{red}{(-1.27)}} \\

\rowcolor{gray!20}
\textit{Qwen2.5 + DCSC (ours.)}
   & 68.37 & \textbf{75.32} & 71.68 & 90.14 {\footnotesize \textcolor{red}{(+6.71)}} & \textbf{84.20 {\footnotesize \textcolor{red}{(+34.20)}}}
   & 84.94 {\footnotesize \textcolor{red}{(+1.51)}} & 53.74 {\footnotesize \textcolor{red}{(+3.74)}} & 0.43 {\footnotesize (+0.43)} & 26.61 {\footnotesize \textcolor{red}{(-2.74)}} & 13.52 {\footnotesize \textcolor{red}{(-1.15)}} \\
\midrule

\rowcolor{gray!20}
\textit{mT5 + DCSC (ours.)}
   & 79.18 & 36.12 & 49.60 & 87.84 {\footnotesize \textcolor{red}{(+4.41)}} & 67.12 {\footnotesize \textcolor{red}{(+17.12)}}
   & \textbf{85.31} {\footnotesize \textcolor{red}{(+1.88)}} & 51.72 {\footnotesize \textcolor{red}{(+1.72)}} & 0.28 {\footnotesize \textcolor{blue}{(+0.28)}} & 27.11 {\footnotesize \textcolor{red}{(-2.24)}} & 13.69 {\footnotesize \textcolor{red}{(-0.98)}} \\

\rowcolor{gray!20}
\textit{KoBart + DCSC (ours.)}
   & \textbf{79.39} & 28.43 & 41.87 & 86.92 {\footnotesize \textcolor{red}{(+3.49)}} & 63.48 {\footnotesize \textcolor{red}{(+13.48)}}
   & 85.00 {\footnotesize \textcolor{red}{(+1.57)}} & 50.33 {\footnotesize \textcolor{red}{(+0.33)}} & 0.23 {\footnotesize \textcolor{blue}{(+0.23)}} & 27.83 {\footnotesize \textcolor{red}{(-1.52)}} & 14.03 {\footnotesize \textcolor{red}{(-0.64)}} \\

\rowcolor{gray!20}
\textit{pkoT5 + DCSC (ours.)$^{\dagger}$}
   & 75.97
   & 70.08
   & \textbf{72.91}
   & \textbf{90.43 {\footnotesize \textcolor{red}{(+7.00)}}}
   & 81.89 {\footnotesize \textcolor{red}{(+31.89)}}
   & 85.16 {\footnotesize \textcolor{red}{(+1.73)}}
   & 53.77 {\footnotesize \textcolor{red}{(+3.77)}}
   & 0.35 {\footnotesize \textcolor{blue}{(+0.35)}}
   & \textbf{26.27 {\footnotesize \textcolor{red}{(-3.08)}}}
   & \textbf{13.30 {\footnotesize \textcolor{red}{(-1.37)}}} \\
\bottomrule

\end{tabular}
}
\vspace{-2mm}
\end{table*}

Table~\ref{tab:main_results} shows the end-to-end performance of \textbf{DCSC} across all corrector backbones. We analyze these results by comparing DCSC with several baselines, and then examine the performance characteristics of different backbone families.

\paragraph{Baseline comparison}
The \textit{Zero-rule} baseline directly accepts the original ASR transcription without any modification. Its \textbf{Acc.} and \textbf{EM} of 83.43 reflect the proportion of naturally error-free utterances in the test set. However, its \textbf{E-WER} remains high at 29.35, showing that erroneous utterances are left entirely uncorrected. This establishes a conservative reference point: an effective post-editing system should reduce \textbf{E-WER} while keeping \textbf{N-WER} low.

The rule-based \textit{IT2-FLS} baseline shows only limited transfer to our Korean STT setting. Although it slightly reduces \textbf{E-WER} and \textbf{Bal-WER}, its detection performance is extremely weak, with a \textbf{Recall} of 2.06 and an \textbf{F1} of 3.84. This suggests that the fuzzy rule-based approach, originally designed for high-WER low-resource ASR outputs, does not transfer effectively to our error-sparse Korean STT setting. The \textit{Char-Seq2Seq} baseline also remains close to the \textit{Zero-rule} baseline, achieving only 7.26 \textbf{Recall} and 12.44 \textbf{F1}, with a marginal \textbf{Bal-WER} improvement. This indicates that a character-level model trained from scratch largely collapses toward identity copying under the highly error-sparse distribution of our dataset. Finally, the direct utterance-to-utterance baselines exhibit severe over-correction. \textit{Llama-3.1+Uttr2Uttr} substantially rewrites clean utterances, leading to an extremely high \textbf{N-WER} and \textbf{Bal-WER}. Although \textit{pkoT5+Uttr2Uttr} is less destructive and slightly reduces \textbf{E-WER}, its \textbf{N-WER} remains high, resulting in a \textbf{Bal-WER} of 25.33. These results show that direct full-sentence generation without an explicit detection gate struggles to preserve already-correct utterances.

\paragraph{Performance overview}
Overall, DCSC consistently outperforms both the passive \textit{Zero-rule} baseline and the active post-editing baselines. All DCSC configurations achieve substantially stronger detection performance than the baselines, with \textbf{F1} scores ranging from 41.87 to 72.91, compared with 30.47 for the best non-proposed baseline. The benefit also carries over to correction quality. All proposed models reduce \textbf{E-WER} below the \textit{Zero-rule} baseline while maintaining very low \textbf{N-WER} values between 0.23 and 0.43, indicating that the correction gate effectively suppresses unnecessary edits on clean utterances. Among all models, \texttt{pkoT5} with DCSC achieves the best correction performance, lowering \textbf{E-WER} from 29.35 to 26.27 and \textbf{Bal-WER} from 14.67 to 13.30. The LLM-based variants also perform strongly: \texttt{Llama-3.1} achieves a \textbf{Bal-WER} of 13.40, while \texttt{Qwen2.5} achieves 13.52. These results demonstrate that the proposed framework improves transcript quality through targeted correction, avoiding the over-correction problem observed in direct utterance-to-utterance generation.

\paragraph{K-Seq2Seq vs. LLMs}
Within DCSC, we observe a clear distinction between the encoder--decoder models and the LLM-based correctors. The LLMs are particularly strong in detection when dialogue context is provided. \texttt{Qwen2.5} achieves the highest \textbf{Recall} of 75.32, while \texttt{Llama-3.1} shows a balanced detection profile with 74.81 \textbf{Precision}, 69.92 \textbf{Recall}, and 72.28 \textbf{F1}. This suggests that LLMs can effectively leverage dialogue-level context to identify potentially erroneous utterances. However, stronger detection does not always translate into the best final correction. The Korean sequence-to-sequence model \texttt{pkoT5} achieves an \textbf{F1} of 72.91 while producing the lowest \textbf{E-WER} and \textbf{Bal-WER}. This indicates that smaller Korean-specialized encoder--decoder models can offer more stable and precise rewriting once the correction decision has been made.

\paragraph{Comparison within K-Seq2Seq models}
Among the encoder--decoder backbones, \texttt{pkoT5} consistently outperforms \texttt{mT5} and \texttt{KoBart}. While \texttt{mT5} and \texttt{KoBart} show high \textbf{Precision} values of 79.18 and 79.39, respectively, their low \textbf{Recall} values of 36.12 and 28.43 indicate that they miss many erroneous utterances. In contrast, \texttt{pkoT5} provides a much better balance between \textbf{Precision} and \textbf{Recall}, resulting in a substantially higher \textbf{F1} of 72.91. This advantage is also reflected in the correction metrics: \texttt{pkoT5} achieves an \textbf{EM} of 85.16 and the highest \textbf{Bal-EM} of 53.77 among the encoder--decoder models, as well as the lowest \textbf{E-WER} of 26.27 and \textbf{Bal-WER} of 13.30. These results suggest that \texttt{pkoT5} is better suited to Korean ASR post-editing than the multilingual \texttt{mT5} and the BART-style Korean backbone.

\paragraph{Comparison within LLMs}
Between the two LLMs, \texttt{Qwen2.5} is more sensitive in detecting errors, achieving the highest \textbf{Recall} of 75.32, an \textbf{F1} of 71.68, and the highest \textbf{Bal-Acc} of 84.20 among all models. However, this sensitivity leads to slightly more unnecessary modification, as reflected in its higher \textbf{N-WER} of 0.43 compared with 0.25 for \texttt{Llama-3.1}. In terms of correcting genuinely erroneous utterances, the two LLMs are very close, with \texttt{Llama-3.1} obtaining an \textbf{E-WER} of 26.56 and \texttt{Qwen2.5} obtaining 26.61. Consequently, \texttt{Llama-3.1} achieves a slightly lower \textbf{Bal-WER} of 13.40 compared with 13.52 for \texttt{Qwen2.5}. These results indicate that \texttt{Qwen2.5} behaves more aggressively in identifying potential errors, whereas \texttt{Llama-3.1} is slightly more conservative and better preserves clean utterances.

\paragraph{Model selection}
Although the LLM-based models show strong detection performance and competitive correction quality, \texttt{pkoT5} with DCSC achieves the best overall correction performance, with an \textbf{EM} of 85.16, the lowest \textbf{E-WER} of 26.27, and the lowest \textbf{Bal-WER} of 13.30. Its detection \textbf{F1} of 72.91 is also higher than that of the \texttt{Qwen2.5} model, despite being substantially smaller and more computationally efficient. Considering this trade-off between effectiveness and deployment cost, we designate \texttt{pkoT5}-based DCSC as the default model for subsequent analyses and as the final proposed framework.

\subsection{Ablation study: Efficacy of token-level detection}
\label{sec:detection_results}

\subsubsection{Setup}
\label{sec:dec_setup}

\paragraph{Evaluation metrics}
 Beyond \textbf{utterance-level} detection metrics introduced in Section~\ref{sec:main_setup}, we assess whether the model can precisely localize errors. Rather than inspecting only error-containing segments, we compare the predicted \textbf{token-level} mask sequence $\hat{M}$ to the corresponding ground-truth $M$ (from Section~\ref{sec:utt2tok}). We compute:
\begin{itemize}
\item \textbf{Exact Match Ratio (EM)}: The fraction of samples for which $\hat{M}$ exactly matches $M$.
\item \textbf{Balanced Exact Match Ratio (Bal-EM)}: \textbf{EM} is computed separately for genuinely normal (\textbf{N-EM}) and erroneous (\textbf{E-EM}) samples, then averaged.
\end{itemize}
These metrics are utilized to measure how precisely the model localizes error spans.

\paragraph{Baselines}
To validate the efficacy of our \textbf{token-level} detection strategy following granularity refinement (Section~\ref{sec:utt2tok}), we establish a baseline by reframing the detection task as a binary classification problem at the \textbf{utterance-level}. In this configuration, the model is tasked with predicting whether a given utterance contains any errors. Specifically, if a transcribed utterance $U$ perfectly matches its ground-truth counterpart $G$, the label $y$ is assigned as 0; otherwise, it is assigned as 1. We fine-tune a detector model for sequence classification using the following weighted loss function:

\begin{equation*}
\begin{split}
\mathcal{L} = -\sum_{i=1}^{N} \Bigl(\;
  & \lambda \, y^{(i)} \, \log\bigl(p(y^{(i)})\bigr) \\
  & + \bigl(1 - y^{(i)}\bigr) \, \log\bigl(1 - p(y^{(i)})\bigr) \Bigr)
\end{split}
\end{equation*}
where $N$ denotes the total number of training samples. For the $i$-th utterance sample $U$, $y^{(i)}$ represents the binary ground-truth label and $p(y^{(i)})$ denotes the predicted probability of the utterance being erroneous (label 1). To mitigate class imbalance, we empirically set the weighting factor $\lambda = 4$.

Furthermore, we incorporate \texttt{KoRoberta}~\citep{NEURIPS_DATASETS_AND_BENCHMARKS_2021_98dce83d}, which has demonstrated stable performance on the KLUE benchmark, as an additional backbone detector. This allows for a comparative analysis against our primary detector, \texttt{KoElectra}.

\subsubsection{Results}
\label{sec:dec_results}

\begin{table*}[t]
\centering
\renewcommand{\arraystretch}{1.2}
\setlength{\tabcolsep}{4pt}
\caption{Comparative analysis between the baseline utterance-level sequence classification and token-level classification methods across two backbone models. Changes relative to the sequence-classification baseline are shown in parentheses, where red indicates performance improvement and blue indicates degradation for each metric. Token-level metrics (\textbf{EM, N-EM, E-EM, Bal-EM}) are only available for \textit{Tok} models.}
\resizebox{0.9\linewidth}{!}{%
\begin{tabular}{l|ccccc|cccc}
\toprule
\multirow{2}{*}{\textbf{Detector}} 
& \multicolumn{5}{c|}{\textbf{Utterance-Level}} 
& \multicolumn{4}{c}{\textbf{Token-Level}} \\
\cmidrule(lr){2-6} \cmidrule(lr){7-10}
& \textbf{Precision} \(\uparrow\) & \textbf{Recall} \(\uparrow\) & \textbf{F1} \(\uparrow\) & \textbf{Acc} \(\uparrow\) & \textbf{Bal-Acc} \(\uparrow\)
  & \textbf{EM} \(\uparrow\) & \textbf{N-EM} \(\uparrow\) & \textbf{E-EM} \(\uparrow\) & \textbf{Bal-EM} \(\uparrow\) \\
\midrule
%------------------------------------------------------------------
\textit{KoRoberta\textsubscript{Seq}}
   & 67.56 & 65.16 & 66.34 & 89.04 & 79.47
   & --  & --  & -- & -- \\
\textit{KoElectra\textsubscript{Seq}}
   & 64.54 & 76.11 & 69.85 & 89.11 & 83.90
   & --  & --  & -- & -- \\
\midrule
\rowcolor{red!10}
\multicolumn{10}{c}{\textit{\textbf{Utterance $\rightarrow$ Tokens}}} \\
\midrule

\textit{KoRoberta\textsubscript{Tok}}
   & 71.35 {\footnotesize \textcolor{red}{(+3.79)}} & 62.11 {\footnotesize \textcolor{blue}{(-3.05)}} & 66.41 {\footnotesize \textcolor{red}{(+0.07)}} & 89.59 {\footnotesize \textcolor{red}{(+0.55)}} & 78.58 {\footnotesize \textcolor{blue}{(-0.89)}}
   & 86.27 & 97.20 & 31.19 & 64.20 \\

\rowcolor{gray!20}
\textit{KoElectra\textsubscript{Tok}}
   & 62.47 {\footnotesize \textcolor{blue}{(-2.07)}} & 81.19 {\footnotesize \textcolor{red}{(+5.08)}} & 70.61 {\footnotesize \textcolor{red}{(+0.76)}} & 88.80 {\footnotesize \textcolor{blue}{(-0.31)}} & 85.75 {\footnotesize \textcolor{red}{(+1.85)}}
   & 84.52 & 94.53 & 34.09 & 64.31 \\
\bottomrule
%------------------------------------------------------------------
\end{tabular}
} 
\vspace{-2mm}
\label{tab:dec_results}
\end{table*}

Table~\ref{tab:dec_results} compares the baseline utterance-level sequence classification (\textit{Seq}) and our proposed token-level classification (\textit{Tok}) methods across two detector backbones. We analyze the results from two perspectives: the effect of supervision granularity and the choice of detector backbone.

\paragraph{Effect of token-level detection}
The \textit{Tok} approach provides a key structural advantage over \textit{Seq} by enabling token-wise error localization, which allows us to evaluate token-level metrics such as \textbf{EM}, \textbf{E-EM}, and \textbf{Bal-EM}. In terms of utterance-level detection, token-level supervision yields modest but consistent gains in \textbf{F1} for both backbones. \texttt{KoRoberta} with \textit{Tok} improves \textbf{F1} from 66.34 to 66.41 and \textbf{Acc} from 89.04 to 89.59, although its \textbf{Recall} and \textbf{Bal-Acc} decrease by 3.05 and 0.89 points, respectively. In contrast, \texttt{KoElectra} benefits more clearly from token-level supervision, increasing \textbf{Recall} from 76.11 to 81.19, \textbf{F1} from 69.85 to 70.61, and \textbf{Bal-Acc} from 83.90 to 85.75. These results suggest that token-level supervision does not simply improve all aggregate metrics uniformly; rather, it changes the detector's behavior by encouraging more fine-grained error localization, with the largest benefit observed for \texttt{KoElectra}.

\paragraph{Model selection: KoElectra vs. KoRoberta}
In our proposed pipeline, the detector functions as a preliminary gate that decides whether an utterance should be forwarded to the downstream corrector. Since the corrector can often recover from false positives by returning clean utterances unchanged, the detector should prioritize minimizing false negatives, making \textbf{Recall} especially important. From this perspective, \texttt{KoElectra} with \textit{Tok} is the most suitable detector, achieving a substantially higher \textbf{Recall} than \texttt{KoRoberta} with \textit{Tok} (81.19 vs. 62.11). It also achieves stronger overall detection performance in \textbf{F1} (70.61 vs. 66.41) and \textbf{Bal-Acc} (85.75 vs. 78.58). At the token level, \texttt{KoElectra} also obtains slightly higher \textbf{E-EM} and \textbf{Bal-EM} scores than \texttt{KoRoberta} (34.09 vs. 31.19 and 64.31 vs. 64.20), indicating better localization of erroneous tokens. We therefore select \texttt{KoElectra} with \textit{Tok} as the default detector for our framework.

%\subsection{Corrector Ablation}
\subsection{Ablation study: Impact of component integration}
\label{sec:correction_results}
We evaluate our error-correction framework and various baselines using the metrics in Section~\ref{sec:main_setup}, progressively adding each proposed component to illustrate its contribution. Table~\ref{tab:cor_results} presents an overview of the findings. We trace the performance trajectory across four refinement stages to isolate the specific contribution of each module.

\subsubsection{Setup}
\label{sec:cor_setup}

\paragraph{Baselines}
To strictly evaluate the contribution of each component in our framework, we conduct a progressive ablation study. We establish a series of baselines starting from coarse-grained \textbf{dialogue-level} correction using LLMs, sequentially integrating our method's components up to a configuration that mirrors our final framework, excluding only the detector module.

First, we fine-tune LLMs on \textbf{dialogue-level} transcriptions, using the entire dialogue context as input and the corresponding ground-truth dialogue as the target. This setup enables the model to correct the full conversation in a single pass. For consistent evaluation, generated corrected dialogues are subsequently split into individual utterances.

Next, following the segmentation approach in Section~\ref{sec:dial2utt}, we adopt an \textbf{utterance-level} baseline. Here, each transcribed utterance serves as a single context, with its corresponding ground truth as the target for fine-tuning both LLMs and K-Seq2Seq models. This configuration aligns with standard post-editing methodologies widely adopted in low-resource language studies~\citep{park-etal-2024-hyper, znotins-gruzitis-2025-conversational}.

Finally, leveraging the refinement described in Section~\ref{sec:utt2span}, we train correctors using \textbf{span-level} target representations. Additionally, following Section~\ref{sec:con_aug}, we establish a context-augmented span-level baseline, denoted as \textit{Con+S}. This configuration serves as the immediate predecessor of DCSC, differing from the full framework only by the absence of detector-guided routing.

\paragraph{Backbone selection}
In this ablation study, we utilized two models as correctors that demonstrated generally superior correction performance in Table~\ref{tab:main_results}. We selected one representative from the K-Seq2Seq family and one from the LLM family, specifically utilizing \texttt{pkoT5} and \texttt{Llama-3.1}, respectively. However, for the entire dialogue-level configuration, we exclusively experimented with \texttt{Llama-3.1} due to the maximum input length constraints inherent in \texttt{pkoT5}.

\begin{table*}[t]
\centering
\renewcommand{\arraystretch}{1.2}
\setlength{\tabcolsep}{4pt}

\caption{
Ablation results averaged over three random seeds. Each result is reported as mean $\pm$ standard deviation. Metrics with $(\uparrow)$ indicate higher is better, whereas those with $(\downarrow)$ indicate lower is better. Boldface denotes the best mean performance for each metric. Significance markers on DCSC indicate paired comparisons against the corresponding strongest non-DCSC baseline, \textit{Con+S}: $^{*}p<0.05$, $^{**}p<0.01$, and $^{***}p<0.001$.
}

\resizebox{\linewidth}{!}{%
\begin{tabular}{l|ccccc|ccccc}
\toprule
\multirow{2}{*}{\textbf{Model}}
  & \multicolumn{5}{c|}{\textbf{Detection}}
  & \multicolumn{5}{c}{\textbf{Correction}} \\
\cmidrule(lr){2-6} \cmidrule(lr){7-11}
  & \textbf{Precision} $(\uparrow)$
  & \textbf{Recall} $(\uparrow)$
  & \textbf{F1} $(\uparrow)$
  & \textbf{Acc} $(\uparrow)$
  & \textbf{Bal-Acc} $(\uparrow)$
  & \textbf{EM} $(\uparrow)$
  & \textbf{Bal-EM} $(\uparrow)$
  & \textbf{N-WER} $(\downarrow)$
  & \textbf{E-WER} $(\downarrow)$
  & \textbf{Bal-WER} $(\downarrow)$ \\
\midrule

\textit{Llama-3.1\textsubscript{D}}
& 15.58 
& 61.13 
& 24.84 
& 38.67 
& 47.67 
& 55.36 
& 37.02 
& 325.36 
& 52.87 
& 189.11  \\
\midrule

\rowcolor{red!10}
\multicolumn{11}{c}{\textit{\textbf{Dialogue $\rightarrow$ Utterances}}} \\
\midrule

\textit{Llama-3.1\textsubscript{U}}
& 10.89 \footnotesize $\pm$ 0.89
& 46.66 \footnotesize $\pm$ 10.09
& 17.66 \footnotesize $\pm$ 0.77
& 39.73 \footnotesize $\pm$ 15.21
& 42.84 \footnotesize $\pm$ 4.02
& 54.33 \footnotesize $\pm$ 16.26
& 32.76 \footnotesize $\pm$ 9.62
& 84.65 \footnotesize $\pm$ 59.60
& 44.46 \footnotesize $\pm$ 13.48
& 64.55 \footnotesize $\pm$ 36.48 \\

\textit{pkoT5\textsubscript{U}}
& 47.41 \footnotesize $\pm$ 20.05
& 30.92 \footnotesize $\pm$ 1.40
& 37.43 \footnotesize $\pm$ 11.94
& 78.43 \footnotesize $\pm$ 9.64
& 57.29 \footnotesize $\pm$ 5.99
& 66.07 \footnotesize $\pm$ 9.51
& 45.35 \footnotesize $\pm$ 5.51
& 15.53 \footnotesize $\pm$ 11.23
& 28.74 \footnotesize $\pm$ 0.42
& 22.13 \footnotesize $\pm$ 5.82 \\

\midrule
\rowcolor{red!10}
\multicolumn{11}{c}{\textit{\textbf{Utterance $\rightarrow$ Error Spans}}} \\
\midrule

\textit{Llama-3.1\textsubscript{S}}
& \textbf{79.87 $\pm$ 3.61}
& 29.35 \footnotesize $\pm$ 2.48
& 42.93 \footnotesize $\pm$ 2.72
& 86.17 \footnotesize $\pm$ 0.40
& 64.21 \footnotesize $\pm$ 1.22
& 82.64 \footnotesize $\pm$ 0.33
& 52.38 \footnotesize $\pm$ 0.13
& 0.42 \footnotesize $\pm$ 0.12
& 29.69 \footnotesize $\pm$ 0.62
& 15.03 \footnotesize $\pm$ 0.26 \\

\textit{pkoT5\textsubscript{S}}
& 65.01 \footnotesize $\pm$ 25.77
& 53.94 \footnotesize $\pm$ 1.70
& 58.96 \footnotesize $\pm$ 16.80
& 83.53 \footnotesize $\pm$ 10.44
& 68.69 \footnotesize $\pm$ 5.64
& 82.94 \footnotesize $\pm$ 10.49
& 52.09 \footnotesize $\pm$ 5.74
& 8.36 \footnotesize $\pm$ 7.51
& 28.49 \footnotesize $\pm$ 0.61
& 15.54 \footnotesize $\pm$ 4.78 \\

\midrule
\rowcolor{red!10}
\multicolumn{11}{c}{\textit{\textbf{+) Context Augmentation}}} \\
\midrule

\textit{Llama-3.1\textsubscript{Con+S}}
& 74.66 \footnotesize $\pm$ 2.55
& 42.30 \footnotesize $\pm$ 2.85
& 54.00 \footnotesize $\pm$ 1.87
& 88.99 \footnotesize $\pm$ 0.62
& 70.38 \footnotesize $\pm$ 1.36
& 83.56 \footnotesize $\pm$ 0.81
& 52.72 \footnotesize $\pm$ 0.73
& 0.46 \footnotesize $\pm$ 0.05
& 26.62 \footnotesize $\pm$ 1.62
& 14.18 \footnotesize $\pm$ 0.81 \\

\textit{pkoT5\textsubscript{Con+S}}
& 77.23 \footnotesize $\pm$ 2.02
& 57.69 \footnotesize $\pm$ 2.45
& 66.05 \footnotesize $\pm$ 1.70
& 90.23 \footnotesize $\pm$ 0.60
& 77.93 \footnotesize $\pm$ 1.26
& 83.52 \footnotesize $\pm$ 0.90
& 53.70 \footnotesize $\pm$ 0.78
& 0.55 \footnotesize $\pm$ 0.07
& 26.48 \footnotesize $\pm$ 1.62
& 13.55 \footnotesize $\pm$ 0.80 \\

\midrule
\rowcolor{red!10}
\multicolumn{11}{c}{\textit{\textbf{+) Detector} (Full DCSC)}} \\
\midrule

\rowcolor{gray!20}
\textit{Llama-3.1\textsubscript{DCSC}}
& 74.81 \footnotesize $\pm$ 7.52
& 69.92$^{***}$ \footnotesize $\pm$ 8.80
& 72.28$^{***}$ \footnotesize $\pm$ 7.08
& 90.07$^{*}$ \footnotesize $\pm$ 0.75
& 80.73$^{***}$ \footnotesize $\pm$ 3.13
& 85.05$^{*}$ \footnotesize $\pm$ 1.44
& \textbf{53.89$^{**}$ \footnotesize $\pm$ 1.22}
& \textbf{0.25$^{***}$ \footnotesize $\pm$ 0.12}
& 26.56 \footnotesize $\pm$ 1.36
& 13.40$^{***}$ \footnotesize $\pm$ 0.89 \\

\rowcolor{gray!20}
\textit{pkoT5\textsubscript{DCSC}}
& 75.97$^{**}$ \footnotesize $\pm$ 6.05
& \textbf{70.08$^{***}$ \footnotesize $\pm$ 5.73}
& \textbf{72.91$^{***}$ \footnotesize $\pm$ 4.91}
& \textbf{90.43 \footnotesize $\pm$ 0.47}
& \textbf{81.89$^{**}$ \footnotesize $\pm$ 2.21}
& \textbf{85.16$^{*}$ \footnotesize $\pm$ 1.25}
& 53.77 \footnotesize $\pm$ 0.59
& 0.35$^{***}$ \footnotesize $\pm$ 0.14
& \textbf{26.27 \footnotesize $\pm$ 1.36}
& \textbf{13.30$^{**}$ \footnotesize $\pm$ 0.85} \\

\bottomrule
\end{tabular}
}

\label{tab:cor_results}
\end{table*}

\subsubsection{Results}
\label{sec:cor_results}

Table~\ref{tab:cor_results} summarizes the ablation results, illustrating how each component progressively contributes to DCSC. We report the mean and standard deviation for every metric to assess both average performance and sensitivity to random initialization. Significance is evaluated using paired predictions on the same test utterances. Detection metrics are tested using a paired bootstrap test, EM and Bal-EM using McNemar's exact test, and WER-based metrics using the Wilcoxon signed-rank test. Each DCSC model is compared with the \textit{Con+S} model using the same backbone.

\paragraph{Dialogue to utterances}
The initial \textbf{dialogue-level} baseline (\textit{D}) exhibits severe hallucination issues. The exceptionally high \textbf{N-WER} of 325.36 and \textbf{E-WER} of 52.87 indicate that the model frequently generates repetitive loops or irrelevant content rather than correcting the transcription. Segmenting the samples into individual \textbf{utterances} (\textit{U}) mitigates this hallucination by shortening the generation scope. For \texttt{Llama-3.1}, \textbf{N-WER} decreases to $84.65 \pm 59.60$; however, its low \textbf{F1} of $17.66 \pm 0.77$ and high \textbf{Bal-WER} of $64.55 \pm 36.48$ show that direct utterance-level generation remains inaccurate and highly variable across seeds. Qualitative inspection suggests that the large variance is mainly caused by occasional output-format failures. In some seeds, the model generates repeated or additional text instead of a single corrected utterance, sharply increasing both the mean and standard deviation of WER. Although \texttt{pkoT5} is less susceptible to these failures, it still records an \textbf{N-WER} of $15.53 \pm 11.23$ and a \textbf{Bal-WER} of $22.13 \pm 5.82$, with substantial variation in \textbf{Precision} and \textbf{F1} ($\sigma=20.05$ and $11.94$). Thus, utterance segmentation alleviates dialogue-level hallucination but does not ensure output-format compliance or reliably determine whether an utterance requires correction.

\paragraph{Utterance to error spans}
The shift to \textbf{span-level} target representations (\textit{S}) marks a critical turning point in performance. For \texttt{Llama-3.1}, \textbf{N-WER} decreases from $84.65 \pm 59.60$ to $0.42 \pm 0.12$, and \textbf{Bal-WER} from $64.55 \pm 36.48$ to $15.03 \pm 0.26$, demonstrating that localized targets suppress both harmful rewriting and seed-dependent instability. \texttt{pkoT5} similarly improves its mean \textbf{F1} from 37.43 to 58.96 and \textbf{Bal-WER} from 22.13 to 15.54. However, its \textbf{Precision}, \textbf{F1}, and \textbf{Bal-WER} remain variable at $65.01 \pm 25.77$, $58.96 \pm 16.80$, and $15.54 \pm 4.78$, respectively. Overall, span-level supervision constrains the output space and enables more precise corrections, although it does not fully stabilize error identification for pkoT5.

\paragraph{Applying context augmentation}
Incorporating dialogue context (\textit{Con+S}) further refines the model's ability to identify and resolve context-dependent errors. For \texttt{Llama-3.1}, context increases mean \textbf{Recall} from 29.35 to 42.30 and \textbf{F1} from 42.93 to 54.00, while reducing \textbf{E-WER} from $29.69 \pm 0.62$ to $26.62 \pm 1.62$ and \textbf{Bal-WER} to $14.18 \pm 0.81$. Dialogue history helps resolve locally plausible but contextually inconsistent errors; for example, a previous mention of ``Eoul Madang Street'' can support recovering an omitted segment when it is later misrecognized as ``Seoul Madang Street''.

The stabilizing effect is especially pronounced for \texttt{pkoT5}: the standard deviations of \textbf{Precision}, \textbf{F1}, \textbf{Acc}, \textbf{EM}, \textbf{N-WER}, and \textbf{Bal-WER} decrease from 25.77, 16.80, 10.44, 10.49, 7.51, and 4.78 to 2.02, 1.70, 0.60, 0.90, 0.07, and 0.80, respectively. Its mean \textbf{F1} concurrently increases from 58.96 to 66.05, while \textbf{Bal-WER} improves from 15.54 to 13.55. Thus, context provides both useful disambiguating information and a more consistent conditioning signal across seeds. After augmentation, both backbones outperform the \textit{Zero-rule} baseline in mean \textbf{Bal-WER}.

\paragraph{Detector integration}
Finally, integrating the detector completes DCSC and yields the strongest overall balance between error coverage and preservation of clean utterances. Relative to \textit{Con+S}, mean \textbf{Recall} increases from 42.30 to 69.92 for \texttt{Llama-3.1} and from 57.69 to 70.08 for \texttt{pkoT5}; mean \textbf{F1} correspondingly increases from 54.00 to 72.28 and from 66.05 to 72.91. Both F1 gains are significant at $p<0.001$, while \textbf{Bal-Acc} rises to 80.73 and 81.89, confirming improved sensitivity to sparse errors.
The detector also better preserves clean utterances: \textbf{N-WER} decreases from $0.46 \pm 0.05$ to $0.25 \pm 0.12$ for \texttt{Llama-3.1} and from $0.55 \pm 0.07$ to $0.35 \pm 0.14$ for \texttt{pkoT5}, with both reductions significant at $p<0.001$. In contrast, \textbf{E-WER} changes only slightly, from 26.62 to 26.56 and from 26.48 to 26.27, indicating that the detector primarily improves routing and suppresses unnecessary edits, while the contextual span corrector determines correction quality on forwarded errors.

Adding a separately trained detector increases routing-related variance: the standard deviations of \textbf{Recall} rise from 2.85 to 8.80 for \texttt{Llama-3.1} and from 2.45 to 5.73 for \texttt{pkoT5}, while those of \textbf{F1} rise from 1.87 to 7.08 and from 1.70 to 4.91. Nevertheless, variability in the principal correction metric increases only slightly, with \textbf{Bal-WER} standard deviation changing from 0.81 to 0.89 and from 0.80 to 0.85. Mean \textbf{Bal-WER} simultaneously improves from 14.18 to 13.40 and from 13.55 to 13.30; paired Wilcoxon tests confirm significance at $p<0.001$ and $p<0.01$, respectively. Therefore, detector integration introduces modest variability but yields statistically reliable average gains. Overall, pkoT5-based DCSC achieves the best results.

%% file: 6_analyses.tex
\section{Analyses}
\label{sec:analyses}
We conduct additional analyses to better understand the behavior and generalizability of our primary model, \texttt{pkoT5}-based DCSC (+ \texttt{Llama-3.1}-based DCSC) within random seed 42. Our analyses proceed from learning dynamics to pipeline behavior and external validity. We then study robustness across speaker roles and error-span counts, followed by qualitative case studies of representative success and failure modes.

\subsection{Representation geometry of input--target formulations}
\label{sec:rep_geometry}

\begin{figure}[!htbp]
\centering

\begin{subfigure}{\columnwidth}
    \centering
    \includegraphics[width=\linewidth]{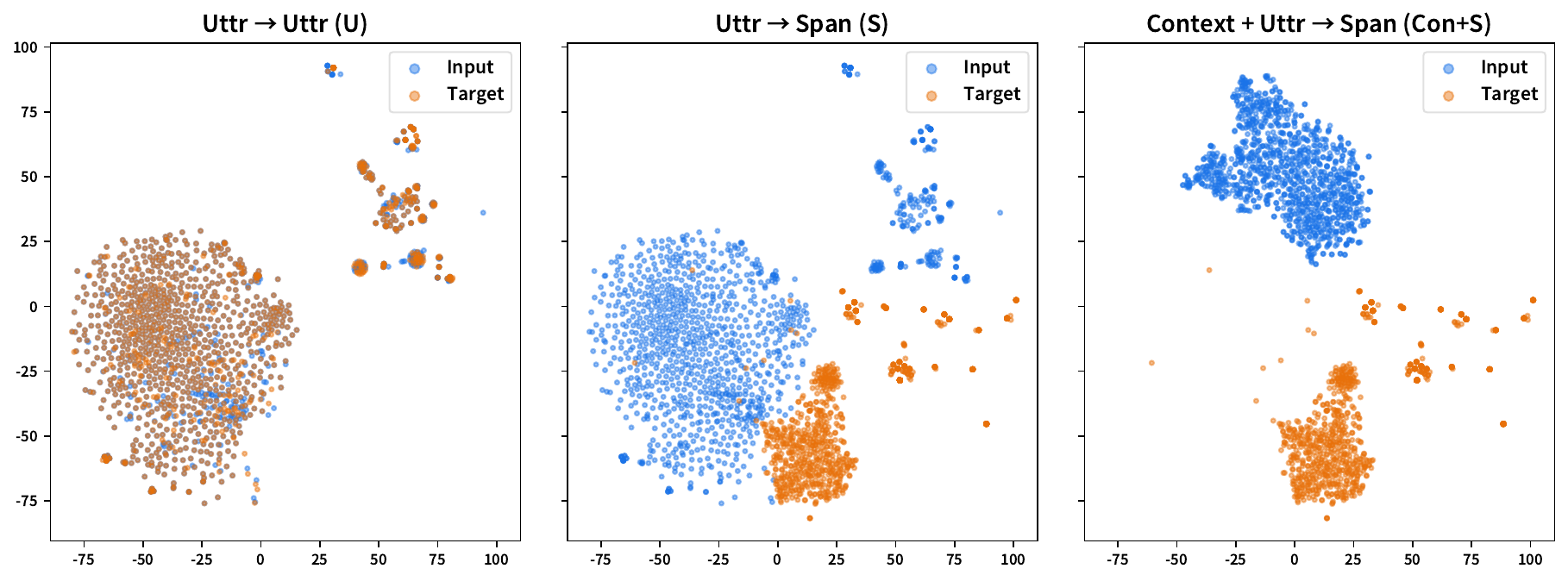}
    \caption{Full error-sparse dataset.}
    \label{fig:input_target_all}
\end{subfigure}

\vspace{0.5em}

\begin{subfigure}{\columnwidth}
    \centering
    \includegraphics[width=\linewidth]{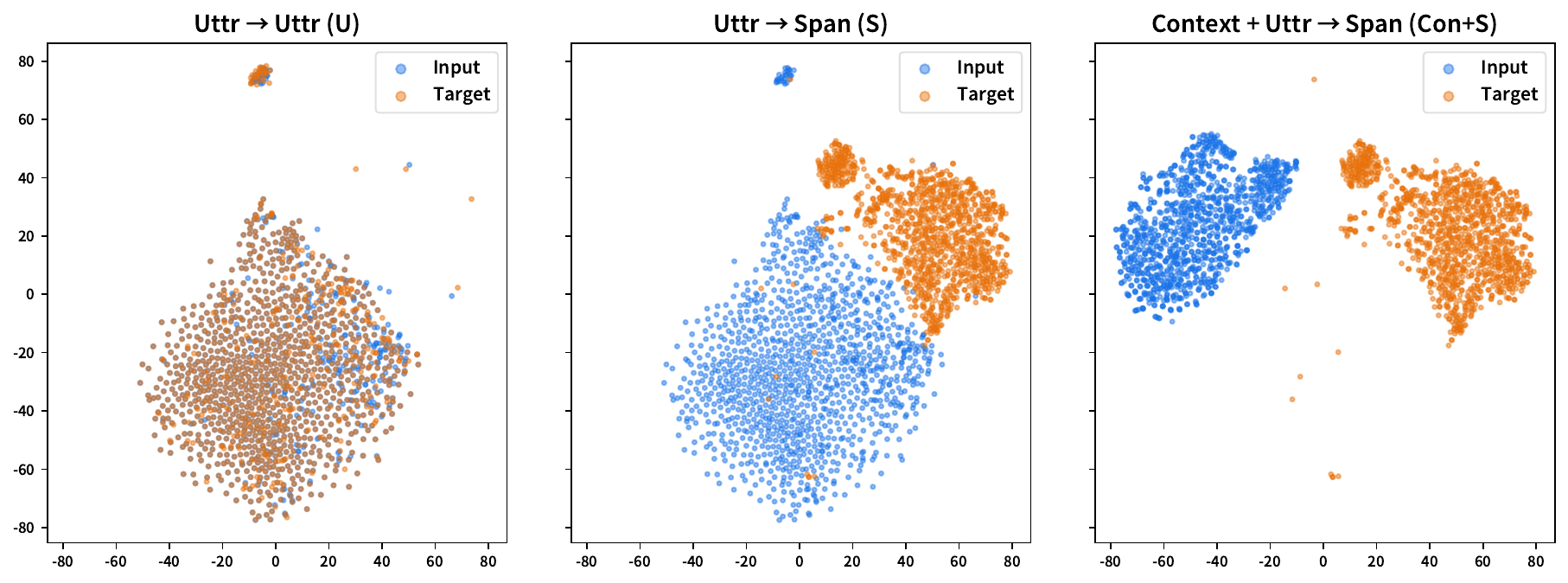}
    \caption{Erroneous utterances only.}
    \label{fig:input_target_error}
\end{subfigure}

\caption{Representation geometry of different input--target formulations. The top panel visualizes the full error-sparse dataset, including error-free samples and their \texttt{No Error} targets, whereas the bottom panel includes only erroneous utterances. From left to right, \textit{U} denotes Uttr$\rightarrow$Uttr, \textit{S} denotes Uttr$\rightarrow$Span, and \textit{Con+S} denotes Context+Uttr$\rightarrow$Span.}
\label{fig:input_target_geometry}
\end{figure}

To qualitatively examine the representation geometry associated with span-level targeting and context augmentation, we use the frozen \texttt{pko-t5-base} encoder to obtain mean-pooled representations of inputs and targets. We compare \textit{U} (Uttr$\rightarrow$Uttr), \textit{S} (Uttr$\rightarrow$Span), and \textit{Con+S} (Context+Uttr$\rightarrow$Span), and project the representations using PCA followed by t-SNE.

Fig.~\ref{fig:input_target_geometry} presents two complementary views. The full-data view reflects the actual error-sparse training distribution, whereas the erroneous-only view removes error-free samples and repeated \texttt{No Error} targets. In both views, \textit{U} shows substantial input--target overlap, consistent with full-reference targets largely reproducing the source utterance. By contrast, the span-level targets in \textit{S} remain more clearly separated from the transcription inputs even after the repeated \texttt{No Error} targets are removed. This suggests that the observed separation is associated not only with the majority-class label but also with the structural difference between full-utterance reconstruction and explicit span correction.

The \textit{Con+S} formulation preserves this target separation while producing a context-enriched input distribution, consistent with dialogue history providing additional information for disambiguation. Nevertheless, because the analysis relies on a frozen encoder and nonlinear dimensionality reduction, it should be interpreted as qualitative and correlational. The controlled ablations in Section~\ref{sec:correction_results} provide the intervention-based evidence for the effects of span-level supervision and context augmentation.

\subsection{Connectivity between detector and corrector}
\label{sec:connection}

\begin{figure}[!htbp]
    \centering
    \includegraphics[width=\columnwidth]{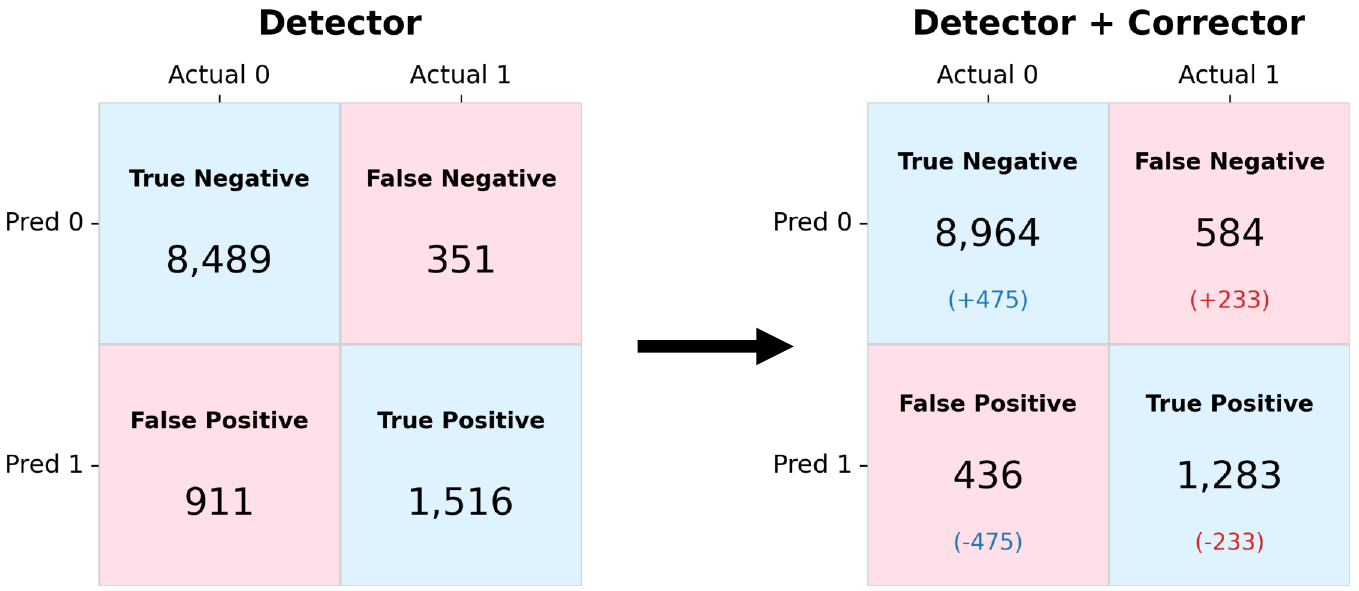}
    \caption{Detector-only and end-to-end confusion matrices. The corrector recovers 52.14\% of detector false positives, improving precision, but leaves 15.37\% of forwarded true errors unresolved, reducing recall.}
    \label{fig:error-propagation}
\end{figure}

\subsubsection{Error propagation}
We conduct a quantitative analysis to examine how detector decisions propagate through the detector--corrector pipeline and how much the corrector can recover from detector misclassifications. Fig.~\ref{fig:error-propagation} compares the detector-only confusion matrix with the end-to-end confusion matrix after the corrector.

First, the detector misses 351 out of 1867 genuinely erroneous samples, corresponding to a false-negative rate of \textbf{18.80\%}. Since these samples are filtered out before reaching the corrector, they constitute an inherent source of under-correction in the pipeline. On the other hand, among the 2427 samples forwarded by the detector as erroneous, 911 are in fact error-free, meaning that \textbf{37.54\%} of the forwarded samples are false positives and therefore vulnerable to over-correction.

Encouragingly, the corrector recovers a substantial portion of these false positives. Specifically, it successfully returns \textbf{475 out of 911} falsely forwarded clean samples unchanged, corresponding to a recovery rate of \textbf{52.14\%}. As a result, the number of false positives is reduced from 911 to 436, which improves \textbf{Precision} from \textbf{62.46} to \textbf{74.64} and decreases \textbf{N-WER} by \textbf{0.16} points.

This robustness, however, comes with a trade-off in recall. Among the 1516 true positives correctly identified by the detector, the corrector fails to resolve \textbf{233} cases, effectively turning them into false negatives. This corresponds to \textbf{15.37\%} of the detector's true positives, increasing the number of false negatives from 351 to 584 and reducing \textbf{Recall} from \textbf{81.20} to \textbf{68.72}. Overall, these results suggest that the corrector behaves as a conservative filter: it substantially mitigates over-correction on clean inputs, but at the cost of leaving some genuine errors unresolved. 
In Section~\ref{sec:threshold}, we examine how threshold calibration changes the number of errors passed to the corrector and quantify the resulting end-to-end precision--recall trade-off.

\subsubsection{Sensitivity to detector thresholds}
\label{sec:threshold}

\begin{figure}[!htbp]
  \centering
  \includegraphics[width=\columnwidth]{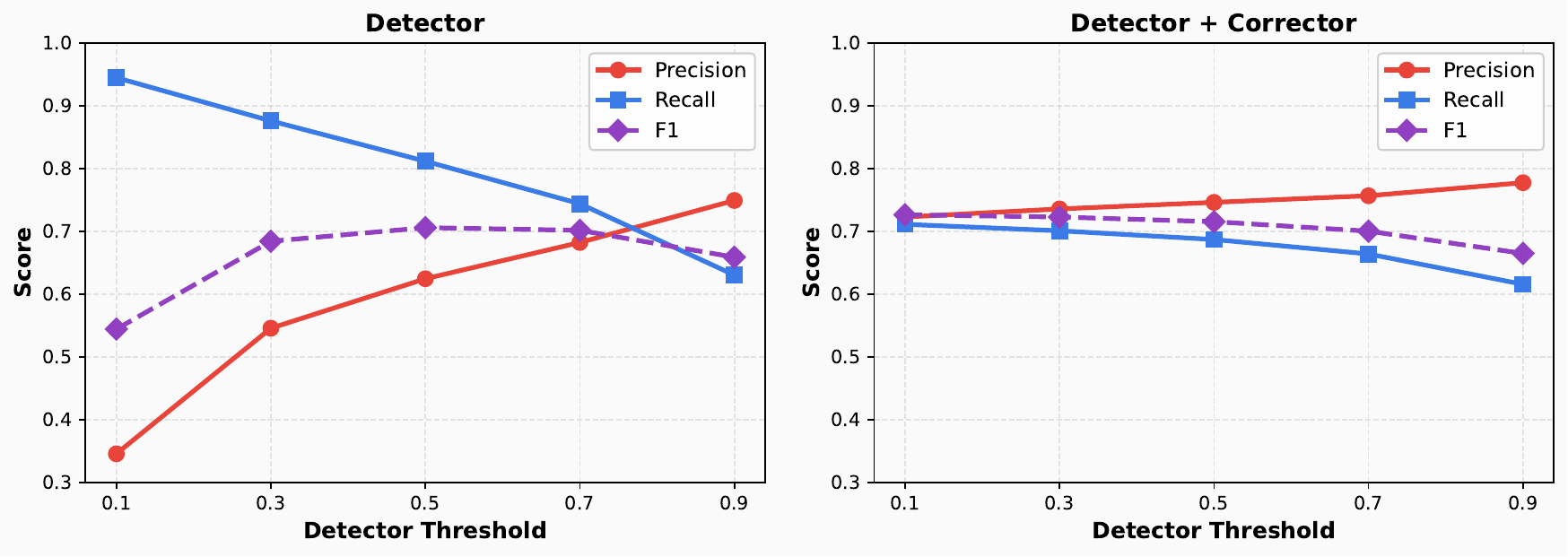}
  \caption{Results obtained by varying the confidence threshold used by the detector to classify samples as erroneous.}
  \label{fig:threshold}
\end{figure}

We further analyze whether the detector-side false negatives observed in Sec.~6.2.1 constitute a fixed structural ceiling or a threshold-dependent operating constraint. We vary the detector confidence threshold over $\{0.1, 0.3, 0.5, 0.7, 0.9\}$ and report detection metrics (\textbf{Precision, Recall, F1}) for both in isolation and after processing through the corrector pipeline in Fig.~\ref{fig:threshold}.

As shown in the left panel, the standalone detector exhibits the expected precision--recall trade-off. Increasing the threshold makes error prediction more conservative, improving \textbf{Precision} while reducing Recall. At the default threshold of 0.5, detector \textbf{Recall} is 81.20\%, yielding a false-negative rate of 18.80\% and filtering out 351 of the 1867 erroneous utterances. In contrast, lowering the threshold to 0.1 raises \textbf{Recall} to approximately 94.5\% and reduces the detector-side false-negative rate to approximately 5.5\%. Thus, the detector-induced upper bound can be substantially relaxed by selecting a recall-oriented threshold.

The right panel shows that the corrector dampens the precision degradation caused by aggressive detector routing. At a threshold of 0.1, standalone detector \textbf{Precision} decreases to approximately 35\% because many clean utterances are forwarded. After the corrector is applied, end-to-end \textbf{Precision} recovers to approximately 72\%, indicating that the corrector successfully returns many falsely forwarded clean utterances unchanged. Accordingly, end-to-end \textbf{F1} varies less sharply than the detector-only metrics and is highest at a relatively low threshold. Nevertheless, the end-to-end \textbf{Recall} at a threshold of 0.1 remains approximately 72\%, despite the detector \textbf{Recall} of approximately 94.5\%. This gap indicates that lowering the threshold effectively mitigates detector-side filtering but does not eliminate corrector-side under-correction. The two stages therefore serve different roles: threshold calibration controls how broadly potential errors are retrieved, whereas the corrector determines whether the forwarded cases are ultimately repaired.

These results demonstrate that the 18.80\% detector false-negative rate at the default threshold is not an immutable structural ceiling. Instead, the framework supports an explicit operating-point choice. Applications in which missed errors are particularly costly can adopt a lower threshold to maximize coverage, while applications with limited correction latency or a higher cost for unnecessary processing may prefer a more conservative threshold. We retain 0.5 as the default setting for the main experiments, while reporting the threshold analysis to make this deployment trade-off explicit.

\subsection{Robustness to autoregressive predicted context}
\label{sec:oracle_context}

During inference, the ground-truth dialogue history used during training is unavailable, so models with context augmentation (\textit{Con+S}, \textit{DCSC}) constructs its context autoregressively from previously corrected utterances $\hat{G}$. To quantify the potential performance loss caused by accumulated correction errors, we compare our \textit{Con+S} setting with an oracle-context setting that provides the preceding ground-truth utterances $G$. The oracle setting serves as an upper bound, while all other experimental conditions remain unchanged.

\begin{table*}[!htbp]
\centering
\small
\caption{Comparison of oracle and predicted inference contexts for the \textit{Con+S} models. Parentheses indicate the change under predicted context relative to the oracle-context upper bound.}
\label{tab:oracle_context}
\resizebox{0.8\linewidth}{!}{%
\begin{tabular}{ll|cccc}
\toprule
\textbf{Backbone}
& \textbf{Inference Context}
& \textbf{F1} $\uparrow$
& \textbf{Bal-Acc} $\uparrow$
& \textbf{Bal-EM} $\uparrow$
& \textbf{Bal-WER} $\downarrow$ \\
\midrule

\multirow{2}{*}{Llama-3.1}
& Oracle $G$
& 55.10
& 70.83
& 54.32
& 13.73 \\
& Predicted $\hat{G}$
& 54.00 {\footnotesize $(-1.10)$}
& 70.38 {\footnotesize $(-0.45)$}
& 52.72 {\footnotesize $(-1.60)$}
& 14.18 {\footnotesize $(+0.45)$} \\
\midrule

\multirow{2}{*}{pkoT5}
& Oracle $G$
& 66.48
& 77.94
& 54.91
& 13.26 \\
& Predicted $\hat{G}$
& 66.05 {\footnotesize $(-0.43)$}
& 77.93 {\footnotesize $(-0.01)$}
& 53.70 {\footnotesize $(-1.21)$}
& 13.55 {\footnotesize $(+0.29)$} \\
\bottomrule
\end{tabular}
}
\end{table*}

As shown in Table~\ref{tab:oracle_context}, the predicted-context setting remains close to the oracle upper bound in detection performance, with \textbf{Bal-Acc} decreasing by only 0.45 points for Llama-3.1 and 0.01 points for pkoT5. The corresponding \textbf{F1} reductions are also limited to 1.10 and 0.43 points. In contrast, the effect is relatively more pronounced on correction quality: predicted context increases \textbf{Bal-WER} by 0.45 points for Llama-3.1 and 0.29 points for pkoT5, while decreasing Bal-EM by 1.60 and 1.21 points, respectively.

These results suggest that using predicted context has little effect on deciding whether an utterance contains an error, but it can slightly affect how the error is corrected. For example, an incorrect word in a previous corrected utterance may make it harder to recover the exact entity or expression in the current utterance. Even so, the \textbf{Bal-WER} increase remains below 0.5 points for both backbones. Therefore, error propagation from predicted context is measurable but limited, and predicted context retains most of the performance obtained with oracle history.

\subsection{Cross-domain transfer}
\label{sec:external_transfer}

\begin{table*}[!htbp]
\centering
\small

\caption{Zero-shot cross-domain evaluation and mixed-domain adaptation results on Hyper-BTS under a controlled error-sparse setting. Changes relative to the Zero-rule baseline are shown in parentheses, where red indicates improvement and blue indicates degradation for each metric.}

\label{tab:hyperbts_transfer}

\resizebox{0.8\linewidth}{!}{%
\begin{tabular}{l|ccccc}
\toprule

\textbf{Model}
& \textbf{EM} $\uparrow$
& \textbf{Bal-EM} $\uparrow$
& \textbf{N-WER} $\downarrow$
& \textbf{E-WER} $\downarrow$
& \textbf{Bal-WER} $\downarrow$ \\

\midrule

\textit{Zero-rule}
& 80.00
& 50.00
& 0.00
& 45.33
& 22.66 \\

\midrule
\midrule

\textit{In-domain SFT (Llama-3.1)}
& 86.91 {\footnotesize \textcolor{red}{(+6.91)}}
& 67.40 {\footnotesize \textcolor{red}{(+17.40)}}
& 0.02 {\footnotesize \textcolor{blue}{(+0.02)}}
& 22.15 {\footnotesize \textcolor{red}{(-23.18)}}
& 11.08 {\footnotesize \textcolor{red}{(-11.58)}} \\

\midrule

\rowcolor{gray!20}
\textit{Out-of-domain DCSC (Llama-3.1)}
& 81.95 {\footnotesize \textcolor{red}{(+1.95)}}
& 52.47 {\footnotesize \textcolor{red}{(+2.47)}}
& 0.84 {\footnotesize \textcolor{blue}{(+0.84)}}
& 35.53 {\footnotesize \textcolor{red}{(-9.80)}}
& 18.19 {\footnotesize \textcolor{red}{(-4.47)}} \\

\rowcolor{gray!12}
\textit{Mixed-domain DSC (Llama-3.1)}
& 86.03 {\footnotesize \textcolor{red}{(+6.03)}}
& 65.07 {\footnotesize \textcolor{red}{(+15.07)}}
& 0.00 {\footnotesize \textcolor{gray}{(+0.00)}}
& 25.83 {\footnotesize \textcolor{red}{(-19.50)}}
& 12.91 {\footnotesize \textcolor{red}{(-9.75)}} \\

\midrule
\midrule

\textit{In-domain SFT (pkoT5)}
& 87.32 {\footnotesize \textcolor{red}{(+7.32)}}
& 68.35 {\footnotesize \textcolor{red}{(+18.35)}}
& 0.00 {\footnotesize \textcolor{gray}{(+0.00)}}
& 21.03 {\footnotesize \textcolor{red}{(-24.30)}}
& 10.52 {\footnotesize \textcolor{red}{(-12.14)}} \\

\midrule

\rowcolor{gray!20}
\textit{Out-of-domain DCSC (pkoT5)}
& 81.35 {\footnotesize \textcolor{red}{(+1.35)}}
& 52.09 {\footnotesize \textcolor{red}{(+2.09)}}
& 1.00 {\footnotesize \textcolor{blue}{(+1.00)}}
& 35.30 {\footnotesize \textcolor{red}{(-10.03)}}
& 18.15 {\footnotesize \textcolor{red}{(-4.51)}} \\

\rowcolor{gray!12}
\textit{Mixed-domain DSC (pkoT5)}
& 86.85 {\footnotesize \textcolor{red}{(+6.85)}}
& 67.13 {\footnotesize \textcolor{red}{(+17.13)}}
& 0.00 {\footnotesize \textcolor{gray}{(+0.00)}}
& 22.72 {\footnotesize \textcolor{red}{(-22.61)}}
& 11.36 {\footnotesize \textcolor{red}{(-11.30)}} \\

\bottomrule
\end{tabular}
}
\end{table*}

To examine cross-domain robustness without conflating zero-shot generalization with domain adaptation, we conduct additional experiments on Hyper-BTS~\citep{park-etal-2024-hyper}, an external Korean text-only ASR post-editing benchmark. Following the error-sparse setting considered throughout this study, we construct controlled Hyper-BTS training and test sets in which 80\% of the inputs are replaced with their corresponding ground-truth sentences, resulting in an 80:20 ratio of error-free to erroneous utterances. This modification is intended to evaluate post-editing behavior under a practically motivated class imbalance; therefore, the reported results should be interpreted specifically under this controlled error-sparse setting rather than as performance on the original Hyper-BTS label distribution.

We evaluate three distinct training conditions. First, \textit{In-domain SFT} is trained solely on the full Hyper-BTS training set (1M size) and serves as an upper-bound reference with direct access to in-domain supervision. Second, \textit{Out-of-domain DCSC} uses a checkpoint trained exclusively on DasanCallDial and is evaluated directly on the Hyper-BTS test set without using any Hyper-BTS examples during training, thereby constituting a genuine zero-shot cross-domain evaluation. Third, \textit{Mixed-domain DSC} is trained on a combination of 50,480 DasanCallDial examples and 50,480 randomly sampled Hyper-BTS examples. The sampled Hyper-BTS subset follows the same 80:20 normal-to-error ratio. Since Hyper-BTS consists of independent utterances rather than dialogue-level interactions, we remove dialogue-context augmentation and refer to the resulting context-free variant of DCSC as \textit{DSC}. This condition evaluates data-efficient domain adaptation rather than zero-shot transfer. All other hyperparameters are kept identical to those used in the main experiments.

As shown in Table~\ref{tab:hyperbts_transfer}, \textit{Out-of-domain DCSC} improves \textbf{Bal-WER} from 22.66 to 18.19 with Llama-3.1 and to 18.15 with pkoT5, despite using no Hyper-BTS training data. The corresponding reductions in E-WER indicate that these gains mainly arise from correcting erroneous utterances, although the slight increases in N-WER show some over-correction under domain shift. \textit{Mixed-domain DSC} further reduces \textbf{Bal-WER} to 12.91 for Llama-3.1 and 11.36 for pkoT5, approaching the \textit{in-domain SFT} results of 11.08 and 10.52, respectively, while using only 50,480 Hyper-BTS examples. The N-WER of 0.00 in both mixed-domain variants further suggests that limited in-domain supervision effectively suppresses over-correction. Overall, the zero-shot results show limited but measurable cross-domain generalization, whereas the mixed-domain results demonstrate data-efficient adaptation toward the in-domain upper bound.

\subsection{Performance by speaker role}
\label{sec:speakers_perform}

\begin{figure}[!htbp]
  \centering
  \includegraphics[width=0.92\linewidth]{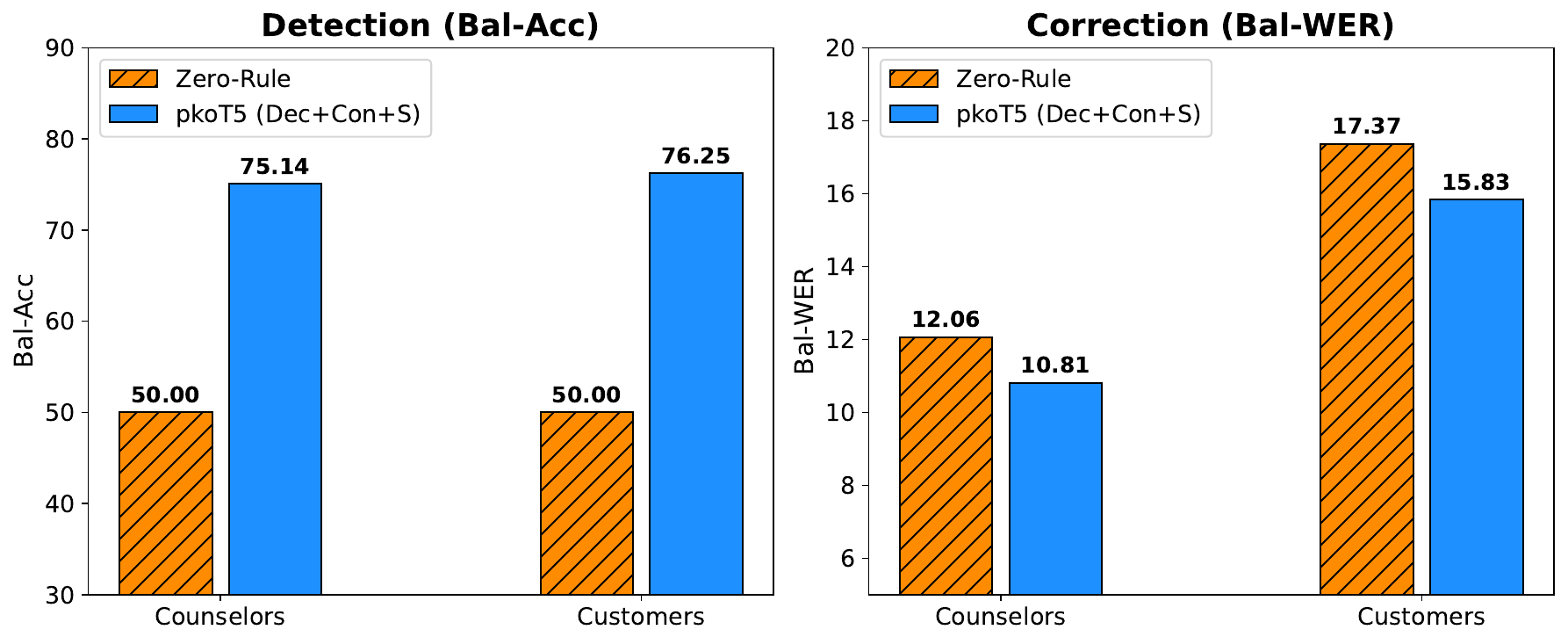}
  \caption{Performance comparison by speaker role.}
  \label{fig:speaker_plot}
\end{figure}

We investigate how our model performs on utterances spoken by call-center counselors versus customers. We report \textbf{Bal-Acc} for detection and \textbf{Bal-WER} for correction, alongside the zero-rule baseline performance, to quantify the practical improvements achieved. Our analysis reveals distinct performance characteristics depending on the speaker's role, influenced by their inherent speech patterns.

As shown in Fig.~\ref{fig:speaker_plot}, regarding \textbf{Bal-Acc}, customer utterances (76.25) yield a slightly higher score than those of counselors (75.14). By contrast, \textbf{Bal-WER} is lower on counselor utterances (10.81) than on customer utterances (15.83). Considering the error rate statistics presented in Section~\ref{sec:speaker_analysis} and \textbf{Bal-WER} of zero-rule baselines in Fig.~\ref{fig:speaker_plot}, this trend likely stems from pronunciation differences. Customer speech is often unstructured and characterized by imprecise pronunciation, leading to diverse misalignments in the transcription. This reduces the likelihood that a token naturally fits the surrounding context, making such errors easier to identify. Conversely, counselor utterances generally consist of structured content with inherently fewer transcription errors, which likely enabled the model to capture more generalized error patterns. Despite the differences between speaker groups, our framework successfully improves both detection and correction performance over the zero-rule baseline for both counselors and customers, thereby reaffirming the robustness of our approach.

\subsection{Performance trend by number of error spans}
\label{sec:num_errors}

\begin{figure*}[!htbp]
  \centering
  \includegraphics[width=0.8\linewidth]{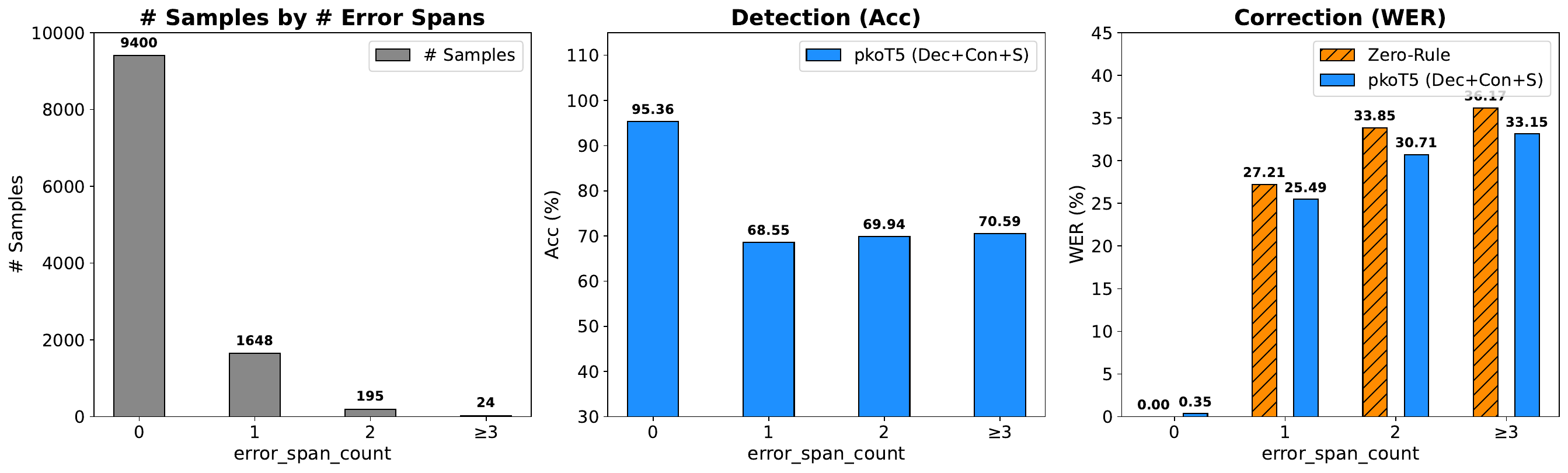}
  \caption{Impact of error-span count. The left histogram shows the distribution of utterances by number of error spans. The middle plot charts detection accuracy, and the right plot charts correction WER: \textbf{N-WER} for error-free samples and \textbf{E-WER} for erroneous ones, alongside the zero-rule baseline.}
  \label{fig:error_span}
\end{figure*}

We further evaluate our model's performance as a function of the number of error spans within each utterance. Given the scarcity of utterances containing four or more error spans, we aggregate them into the subset of three or more ($\geq 3$) spans. The left panel of Fig.~\ref{fig:error_span} illustrates the highly imbalanced distribution: the test set contains 9400 error-free utterances, while 1648 of the 1867 erroneous utterances contain only a single error span. Thus, single-span utterances account for approximately 88\% of all erroneous samples, whereas utterances with two and three or more spans comprise only 195 and 24 samples, respectively.

The middle panel of Fig.~\ref{fig:error_span} reports utterance-level detection Accuracy (\textbf{Acc}). For error-free utterances, the model achieves an \textbf{Acc} of 95.36\%, indicating that most clean inputs are correctly preserved. For erroneous utterances, detection performance remains relatively consistent across different error densities: \textbf{Acc} is 68.55\% for one span, 69.94\% for two spans, and 70.59\% for three or more spans. Although additional error spans provide slightly stronger evidence for detection, the small differences across the erroneous subsets suggest that the model's detection capability is not strongly dependent on error density. In particular, the model maintains comparable performance for the single-span subset, which represents the dominant error condition in the dataset.

The right panel of Fig.~\ref{fig:error_span} presents correction performance in terms of \textbf{WER}, with the zero-rule baseline included for comparison. For error-free utterances, the model yields an \textbf{N-WER} of only 0.35, demonstrating that the majority of clean text remains unchanged despite occasional false-positive routing. For every erroneous subset, the proposed model reduces \textbf{E-WER} relative to the zero-rule baseline. Specifically, \textbf{E-WER} decreases from 27.21 to 25.49 for single-span utterances, from 33.85 to 30.71 for two-span utterances, and from 36.17 to 33.15 for utterances with three or more spans. The larger absolute reductions observed for multi-span utterances indicate that the model can correct multiple errors when stronger error evidence is available. At the same time, the improvement on the substantially larger single-span subset confirms that the overall correction gains are not driven solely by the relatively rare high-error-density cases. Overall, DCSC consistently improves erroneous utterances across different levels of error complexity while introducing only limited changes to error-free inputs.

\begin{figure*}[!htbp]
  \centering
  \includegraphics[width=1\linewidth]{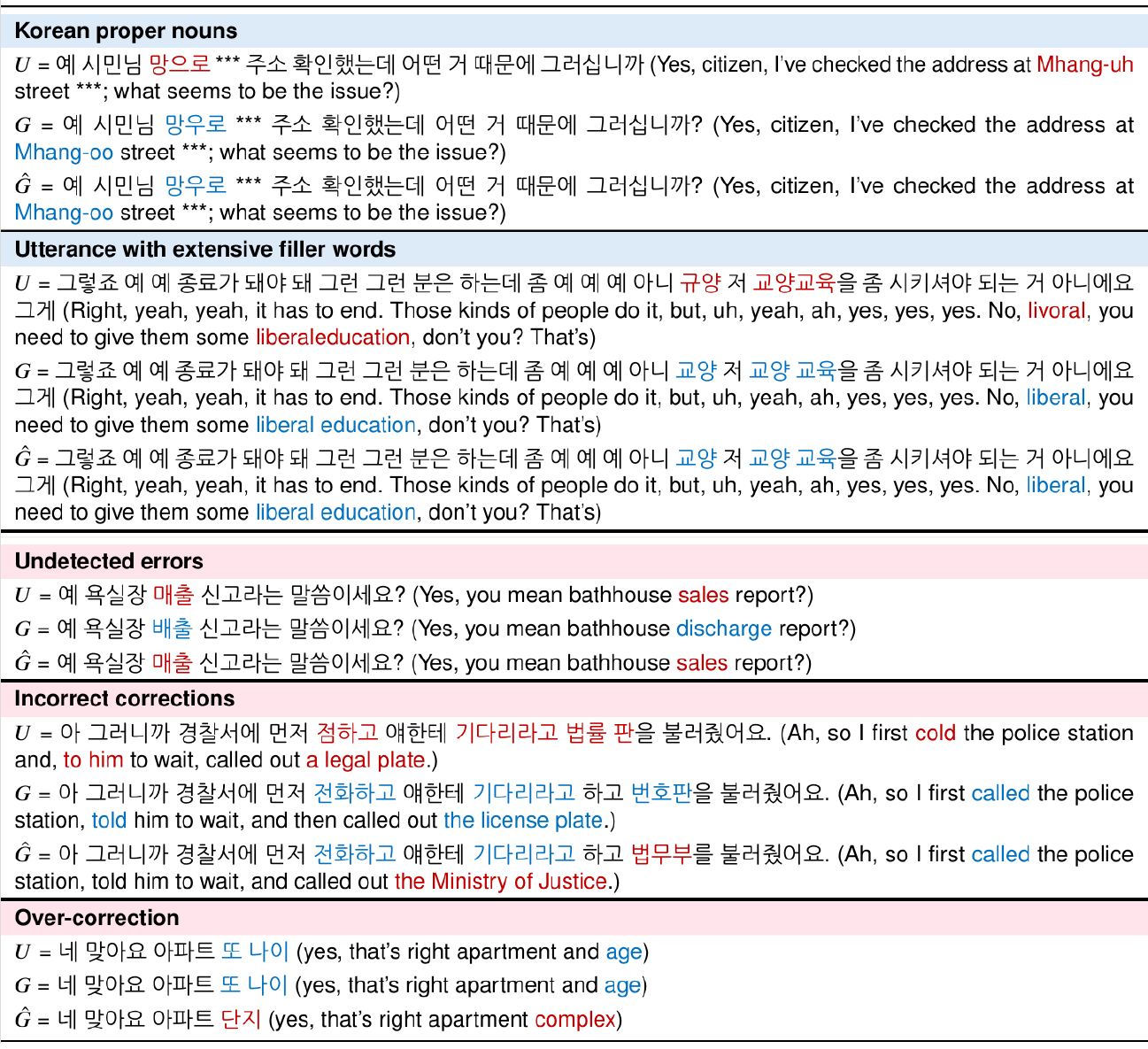}
  \caption{Case-study examples. The first two blocks (light blue background) illustrate scenarios where our corrector succeeds. The remaining three blocks (light red) highlight failure modes. Within each row, erroneous tokens are marked in \textcolor{MyRed}{red} and their correct counterparts in \textcolor{MyBlue}{blue}.}
\label{fig:case_study}
\end{figure*}

\subsection{Case study}
\label{sec:case_study}

We conduct a qualitative analysis of the outputs generated by our model. As summarized in Fig.~\ref{fig:case_study}, we identify two representative \emph{success} scenarios and three representative \emph{failure} scenarios.

\paragraph{Success cases} \textbf{(1) Korean proper nouns.} Our model robustly fixes mistranscribed Korean-specific proper nouns, such as addresses, institutions, and event names. In the first example of Fig.~\ref{fig:case_study}, the erroneous ``Mahng-uh ro'' is accurately corrected to ``Mahng-oo ro (Mahng-oo Street)''. Such corrections are difficult unless the backbone model possesses specialized knowledge of Korean toponyms, underscoring the advantage of adopting a Korean-pretrained model over a general-purpose LLM.\newline
\textbf{(2) Utterances with extensive filler words.} Our model remains stable even when the input contains heavy verbal padding. The second example in Fig.~\ref{fig:case_study} shows an utterance cluttered with fillers, yet our framework successfully isolates and amends the true error spans. This resilience stems from the finer granularity of both inputs and targets, which allows the model to ignore meaningless tokens and focus on genuinely erroneous segments.

\paragraph{Failure cases} \textbf{(1) Undetected errors.} When the prepended detector misclassifies an erroneous utterance as error-free, the corrector is never invoked. As illustrated in the third example, the detector overlooks the misrecognized word ``maechul (sales)'' because the utterance appears locally fluent; identifying the error would require broader dialogue context.\newline
\textbf{(2) Incorrect corrections.} In certain customer utterances, our model fails to propose an appropriate fix even after the detector flags an error. These cases typically involve severely fragmented or context-dependent speech, making it difficult to infer the intended form without additional audio or dialogue cues. The fourth example highlights such a scenario.\newline
\textbf{(3) Over-correction of fluent utterances.} Occasionally, the two-stage pipeline mistakenly presumes that a fluent utterance is erroneous and generates an alternative form. The fifth example shows an utterance whose original version is arguably acceptable, yet our model produces a more coherent variant. Although this technically constitutes an error, the resulting sentence may be more fluent than the ground truth.

%% file: 7_discussion.tex
\section{Discussion}
\subsection{Potential limitations and future directions}

Our study primarily focuses on post-editing transcripts produced by a Korean ASR system; however, the proposed methodology holds potential for extension to other languages and tasks. Evaluating whether the span-based granularity and the detector-corrector design yield similar benefits in other low-resource languages remains an important avenue for future work. Furthermore, while our experiments assume a text-only setting reflecting scenarios where raw audio is inaccessible, richer multimodal approaches could be explored when audio data is available.

\paragraph{Dataset generalizability and industrial impact}
Since our proposed \textbf{DasanCallDial} is derived from actual call center operations, it inherently exhibits characteristics specific to that domain, such as standardized response protocols and frequent turn-taking. This creates a potential risk that the model may not generalize perfectly to open-domain conversations. However, as demonstrated by the topic and frequency analyses in Sections~\ref{sec:topic} and \ref{sec:frequency}, \textbf{DasanCallDial} encompasses a substantial diversity of subjects and expressions, indicating a degree of task transferability. Beyond its academic value, our proposed framework offers significant industrial implications. By validating the feasibility of effective post-editing in real-world, high-noise environments without requiring audio, our work provides a practical blueprint for enhancing automated customer service systems, CRM data analysis, and other downstream applications where transcription accuracy is paramount but resource constraints exist.

\paragraph{Interface granularity and optimization}
A key limitation of the current pipeline is the coarse interface between the detector and the corrector. Although the detector predicts token-level masks, the decision to invoke the corrector is made at the utterance level. Directly passing token-level error indices to the corrector could potentially yield more targeted edits; however, this approach risks propagating detection errors and increasing the model's sensitivity to span boundary precision. Moreover, such fine-grained integration necessitates complex input engineering, which may hinder optimization stability. We leave the exploration of this tighter integration for future research.

\paragraph{Addressing under-correction}
The gated pipeline remains susceptible to under-correction from two sources. First, erroneous utterances may be filtered out by the detector before reaching the corrector. Second, even forwarded utterances may remain unresolved when the corrector returns the original input or produces an ineffective edit. As demonstrated in Section~\ref{sec:threshold}, the first source can be substantially reduced by lowering the detector threshold, although this increases the number of clean utterances sent for correction. The corrector mitigates part of this cost by filtering false-positive routing decisions, but threshold calibration does not address corrector-side failures. Future work should therefore investigate improved confidence calibration and training objectives that more directly penalize unresolved erroneous utterances.

\paragraph{Contextual limitations}
As discussed in Section~\ref{sec:case_study}, adopting finer granularity reduces the immediate context available to the model. This potentially limits the correction of errors that depend on long-term dialogue cues not present in the current input window. Future studies might address this challenge by incorporating compressed dialogue-context embeddings or developing more sophisticated context augmentation strategies to capture broader discourse dependencies.

\paragraph{Metadata and personalization}
Finally, although our dataset captures speech from a diverse user base, it currently lacks explicit meta-information about the speakers. Even minimal persona descriptors, such as regional background or age group, could enable personalized correction strategies, including dialect-aware editing. Future efforts could focus on integrating such metadata—while strictly adhering to privacy protocols—to further enhance correction accuracy and adaptability.

\subsection{Ethical and copyright considerations}
Ethical considerations were carefully managed throughout this study, which involved analyzing real-world call-center data provided by \textbf{Dasan Call Foundation}, an official government organization in Korea. The creation of ground-truth annotations and the de-identification process were carried out securely by two trained Call Foundation employees with direct audio access (details in Section~\ref{sec:dataset_construction}). Only the anonymized dataset (final \textbf{DasanCallDial} dataset) was shared with \textbf{Chung-Ang University} strictly for research purposes.

Regarding copyright, \textbf{DasanCallDial} derived from authentic call transcripts complies with legal standards for academic research, provided it is non-commercial. Our study adheres to these regulations, explicitly limiting the dataset's usage to scholarly research aimed at advancing language model development, without intent for distribution or commercial exploitation. Furthermore, to ensure the dataset is utilized strictly for research purposes while restricting commercial exploitation, we release the \textbf{DasanCallDial} dataset under the \textbf{CC BY-NC 4.0} license. Additionally, we provide \textbf{full access} to the source code for training our primary model (pkoT5-based DCSC) under an open-source license.

% You can see the dataset and the source code at \url{https://github.com/imsongpasimin/Dasan_post-editing.git}.

%% file: 8_conclusion.tex
\section{Conclusion}\label{sec:conclusion}
In this work, we introduced \textbf{DasanCallDial}, the first large-scale real-world benchmark for Korean ASR error correction, comprising 1974 call-center dialogues with 115,460 utterance lines with manually verified references.  Utilizing this resource, we developed Detector-Gated Contextual Span Correction (DCSC), a robust two-stage framework that combines token-level detection, detector-guided routing, dialogue-context augmentation, and span-level correction. Our primary model, integrating an encoder-based KoElectra detector with a context-aware pkoT5 corrector, achieves state-of-the-art performance, significantly outperforming both zero-rule baselines and general-purpose LLMs, in a label-imbalanced setting. Extensive experiments characterizing the learning dynamics, transferability, and robustness validate the effectiveness of DCSC in addressing the distinctive challenges of Korean ASR output. Collectively, our contributions provide a foundational dataset and a strong baseline, paving the way for future advancements in Korean post-editing and low-resource speech technologies.

%% file: 9_appendix.tex
\clearpage
\appendix
%% Keep figure/table numbering continuous across the appendices, as in the
%% published version (Fig. 12-14, Table 9-11) instead of elsarticle's A.1 style.
\makeatletter\gdef\thefigure{\arabic{figure}}\gdef\thetable{\arabic{table}}\makeatother

\section{Dataset examples}\label{app:details}
\begin{figure*}[htbp]
  \centering
  \includegraphics[width=\linewidth]{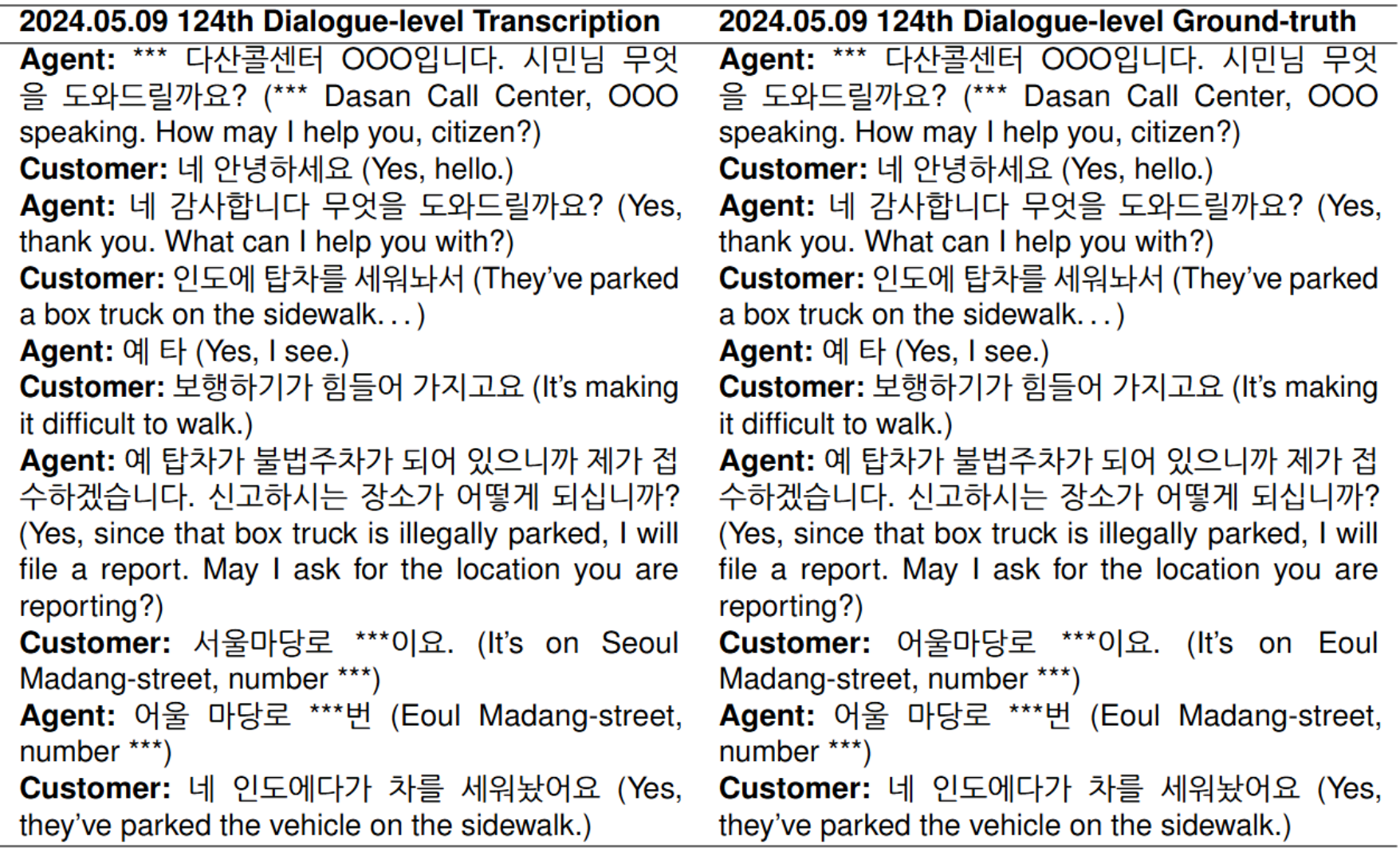}
    \caption{Dialogue-level Dataset: 2024.05.09 124th Call}
    \label{fig:before}
\end{figure*}

\begin{figure*}[htbp]
  \centering
  \includegraphics[width=\linewidth]{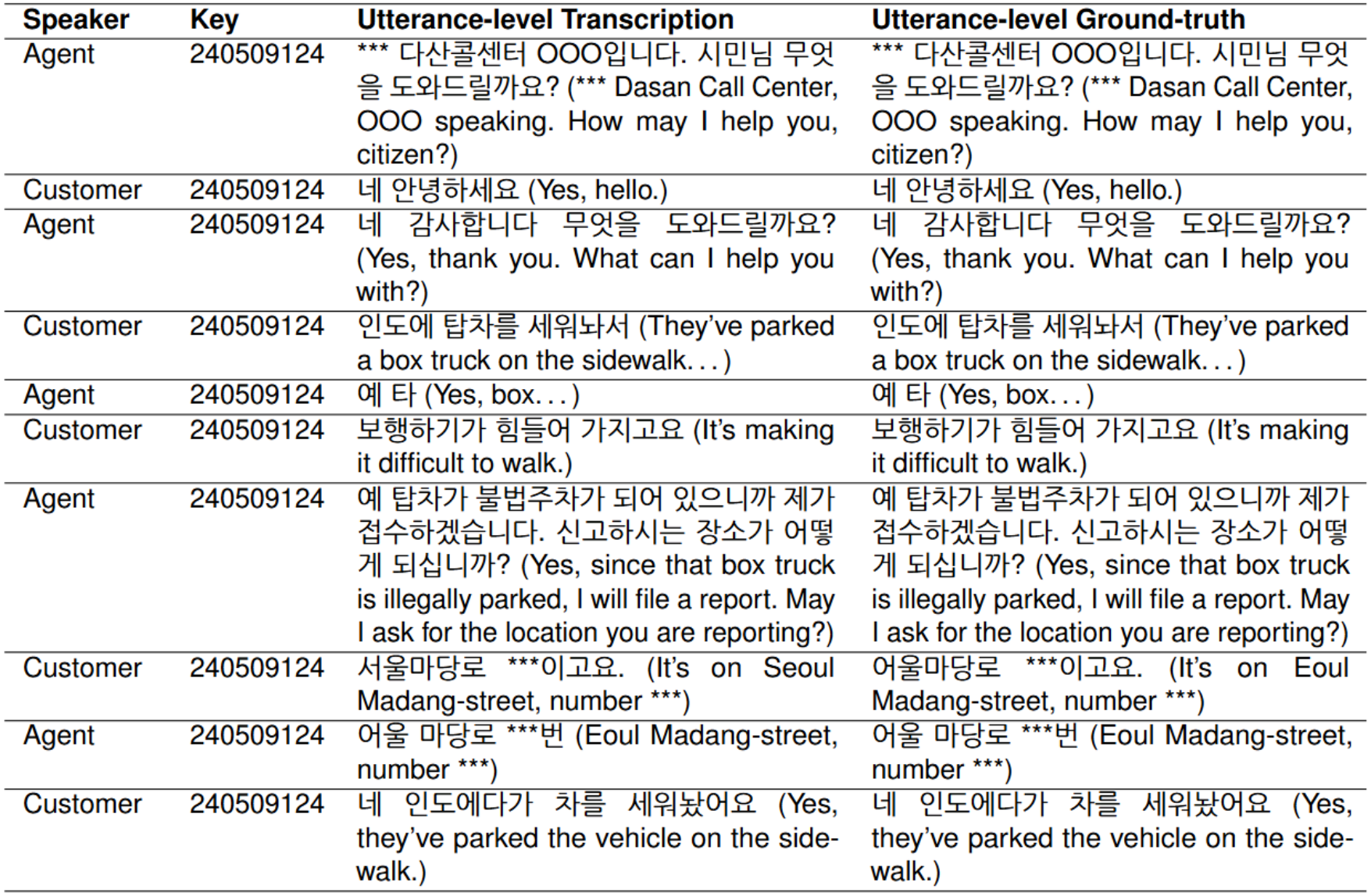}
    \caption{Utterance-level Dataset: 2024.05.09 124th Call}
    \label{fig:after}
\end{figure*}

This section presents concrete examples of the original \textbf{dialogue-level} dataset (Fig.~\ref{fig:before}) alongside the \textbf{utterance-level} version (Fig.~\ref{fig:after}). The side-by-side tables illustrate the dataset's structure, typical content, and the outcome of our de-identification process.

\section{Annotation guideline}
\label{app:anno_guide}

Table~\ref{tab:annotation_guidelines} provides the detailed annotation guideline in Section~\ref{sec:dial_data}.

\begin{table}[t]
\centering
\small
\caption{Annotation guidelines for constructing ground-truth transcriptions.}
\label{tab:annotation_guidelines}
\begin{tabularx}{\linewidth}{|p{0.34\linewidth}|X|}
\hline
\textbf{Case} & \textbf{Annotation guideline} \\
\hline
Clear phonetic recognition error
& Correct the transcription according to the audio signal and the surrounding dialogue context. \\
\hline
Speaker's actual ungrammatical expression
& Preserve the original utterance without grammatical normalization if the expression was actually spoken by the speaker. \\
\hline
Fillers, pauses, and repetitions
& Preserve fillers, pauses, and repeated expressions when they are present in the actual speech. \\
\hline
Numbers, addresses, and phone numbers
& Keep masking according to the privacy protection rules. \\
\hline
Domain-specific proper nouns
& Correct them based on the consultation context and standardized institutional terminology. \\
\hline
Ambiguous phonetic recognition
& Determine the correction using the surrounding dialogue context and consultation-domain knowledge; if the intended expression remains uncertain, preserve the original transcription conservatively. \\
\hline
Homophones and spacing variants
& Apply Korean orthographic conventions while preserving semantic consistency with the utterance and dialogue context. \\
\hline
\end{tabularx}
\end{table}

\section{Cross-ASR error distribution analysis}
\label{app:cross_asr}

To examine whether the error-sparse characteristics of DasanCallDial are specific to \texttt{HAIV}, we additionally transcribe the raw call-center audio using a Korean-fine-tuned \texttt{Whisper} model (ghost613/Whisper-large-v3-turbo-korean) and evaluate it against the same human-annotated references. \texttt{Whisper} produces an utterance error rate of 20.90\%, compared with 17.95\% for \texttt{HAIV}. Although \texttt{Whisper} exhibits a moderately higher error rate, approximately four out of five utterances remain error-free, preserving the clean-majority distribution observed with \texttt{HAIV}.

\begin{figure*}[h]
\centering
\includegraphics[width=\linewidth]{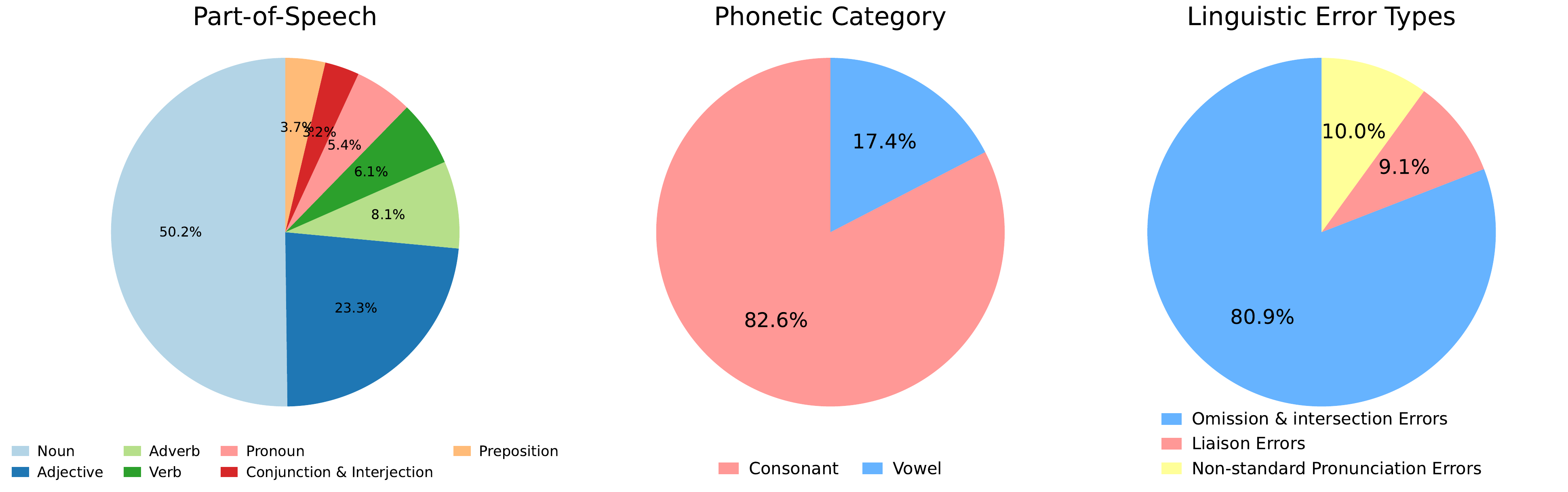}
\caption{Error distributions obtained by re-transcribing the DasanCallDial audio with a Korean-fine-tuned \texttt{Whisper} model. Errors are categorized by part of speech, phonetic category, and linguistic error type following the same criteria used for \texttt{HAIV} in Fig.~\ref{fig:error_type}.}
\label{fig:Whisper_error_analysis}
\end{figure*}

As shown in Fig.~\ref{fig:Whisper_error_analysis}, the major error patterns closely align with those observed for \texttt{HAIV} in Fig.~\ref{fig:error_type}. Nouns remain the most affected part of speech (50.2\% for \texttt{Whisper} versus 56.3\% for \texttt{HAIV}), while consonant-related errors remain dominant (82.6\% versus 84.5\%). Omission and insertion errors are also the most frequent linguistic error type (80.9\% versus 75.2\%), and the proportions of non-standard pronunciation errors are nearly identical (10.0\% versus 10.2\%). Differences in secondary part-of-speech categories and liaison errors indicate some recognizer-specific variation, but the overall failure-mode structure remains consistent across the two ASR systems.

The moderately lower error rate of \texttt{HAIV} may partly reflect that the reference transcripts were originally constructed by manually post-editing \texttt{HAIV} outputs, which can favor its segmentation and transcription conventions. Nevertheless, the small error-rate gap and the consistent dominant error patterns suggest that the error-sparse setting and principal error characteristics are not artifacts unique to \texttt{HAIV}.

\section{Detailed experimental setup}
\label{appendix:exp_setup}

Tables~\ref{tab:appendix_model_resources} and~\ref{tab:appendix_training_config} provide the detailed experimental setup corresponding to Section~\ref{sec:main_setup}. Table~\ref{tab:appendix_model_resources} summarizes the backbone architectures and computational requirements, while Table~\ref{tab:appendix_training_config} reports the training and inference configurations for each model family.

\begin{table*}[t]
\centering
\footnotesize
\caption{Backbone-specific architecture and computational requirements.}
\label{tab:appendix_model_resources}
\begin{tabularx}{\linewidth}{|p{0.22\linewidth}|p{0.20\linewidth}|X|}
\hline
\textbf{Backbone} & \textbf{Item} & \textbf{Details} \\
\hline
KoElectra detector & Architecture & 12-layer ELECTRA encoder for binary token classification; hidden size 768; 12 attention heads; approximately 110M parameters. \\
KoElectra detector & Training duration & Approximately 1 h 30 min. \\
KoElectra detector & Inference latency & Approximately 0.5--1.6 ms per utterance. \\
KoElectra detector & Required GPU & 1 NVIDIA RTX 6000 Ada. \\
\hline
pkoT5 corrector & Architecture & Encoder--decoder Transformer with 12 encoder and 12 decoder layers; hidden size 768; 12 attention heads; approximately 220M parameters. \\
pkoT5 corrector & Training duration & Approximately 2 h 20 min. \\
pkoT5 corrector & Inference latency & Approximately 36 ms per utterance. \\
pkoT5 corrector & Required GPU & 1 NVIDIA RTX 6000 Ada. \\
\hline
Llama-3.1 corrector & Architecture & Decoder-only Transformer with 32 layers; hidden size 4096; grouped-query attention; approximately 8B parameters; fine-tuned with LoRA. \\
Llama-3.1 corrector & Training duration & Approximately 7 h 20 min. \\
Llama-3.1 corrector & Inference latency & Approximately 2.5 s per utterance. \\
Llama-3.1 corrector & Required GPU & 2 NVIDIA RTX 6000 Ada GPUs. \\
\hline
\end{tabularx}
\end{table*}

\begin{table*}[t]
\centering
\footnotesize
\caption{Backbone-specific training and inference configurations.}
\label{tab:appendix_training_config}
\begin{tabularx}{\linewidth}{|p{0.22\linewidth}|p{0.20\linewidth}|X|}
\hline
\textbf{Backbone} & \textbf{Item} & \textbf{Details} \\
\hline
KoElectra detector & Training configuration & AdamW optimizer; learning rate \texttt{2e-5}; batch size 24; 15 epochs; maximum sequence length 128. \\
KoElectra detector & Input and output & The input is an utterance-level ASR transcription, and the output is a binary token-level error label sequence. \\
KoElectra detector & Checkpoint selection & The best checkpoint is selected based on validation F1. \\
KoElectra detector & Inference & An utterance is forwarded to the corrector if at least one token is predicted as erroneous. \\
\hline
pkoT5 corrector & Training configuration & AdamW optimizer; learning rate \texttt{1e-4}; batch size 24; weight decay \texttt{1e-3}; 15 epochs. \\
pkoT5 corrector & Input and output & The maximum source length is 512 and the maximum target length is 128. The input consists of the target utterance with dialogue context, and the target is the span-level correction string. \\
pkoT5 corrector & Checkpoint selection & The best checkpoint is selected based on validation Bal-WER. \\
pkoT5 corrector & Inference & Greedy decoding is used during inference. \\
\hline
Llama-3.1 corrector & Training configuration & LoRA fine-tuning with DeepSpeed ZeRO-3; AdamW optimizer; learning rate \texttt{1e-4}; batch size 24; gradient accumulation step 8; 6 epochs. \\
Llama-3.1 corrector & Input and output & Decoder-only instruction-tuning format. The prompt contains the target utterance and dialogue context, and the target is the span-level correction string. \\
Llama-3.1 corrector & Checkpoint selection & The best checkpoint is selected based on validation Bal-WER. \\
Llama-3.1 corrector & Inference & Greedy decoding is used during inference. \\
\hline
\end{tabularx}
\end{table*}

%% file: 10_ref.bib
@ARTICLE{10301513,
  author={Prabhavalkar, Rohit and Hori, Takaaki and Sainath, Tara N. and Schlüter, Ralf and Watanabe, Shinji},
  journal={IEEE/ACM Transactions on Audio, Speech, and Language Processing}, 
  title={End-to-End Speech Recognition: A Survey}, 
  year={2024},
  volume={32},
  number={},
  pages={325-351},
  doi={10.1109/TASLP.2023.3328283}}

@article{AHLAWAT2025201,
title = {Automatic Speech Recognition: A survey of deep learning techniques and approaches},
journal = {International Journal of Cognitive Computing in Engineering},
volume = {6},
pages = {201-237},
year = {2025},
issn = {2666-3074},
doi = {https://doi.org/10.1016/j.ijcce.2024.12.007},
url = {https://www.sciencedirect.com/science/article/pii/S2666307424000573},
author = {Harsh Ahlawat and Naveen Aggarwal and Deepti Gupta}
}

@inproceedings{10.5555/3495724.3496768,
author = {Baevski, Alexei and Zhou, Henry and Mohamed, Abdelrahman and Auli, Michael},
title = {wav2vec 2.0: a framework for self-supervised learning of speech representations},
year = {2020},
isbn = {9781713829546},
publisher = {Curran Associates Inc.},
address = {Red Hook, NY, USA},
booktitle = {Proceedings of the 34th International Conference on Neural Information Processing Systems},
articleno = {1044},
numpages = {12},
location = {Vancouver, BC, Canada},
series = {NIPS '20}
}

@inproceedings{gulati20_interspeech,
  title     = {{Conformer: Convolution-augmented Transformer for Speech Recognition}},
  author    = {Anmol Gulati and James Qin and Chung-Cheng Chiu and Niki Parmar and Yu Zhang and Jiahui Yu and Wei Han and Shibo Wang and Zhengdong Zhang and Yonghui Wu and Ruoming Pang},
  year      = {2020},
  booktitle = {{Interspeech 2020}},
  pages     = {5036--5040},
  doi       = {10.21437/Interspeech.2020-3015},
  issn      = {2958-1796},
}

@INPROCEEDINGS{9053051,
  author={Hrinchuk, Oleksii and Popova, Mariya and Ginsburg, Boris},
  booktitle={ICASSP 2020 - 2020 IEEE International Conference on Acoustics, Speech and Signal Processing (ICASSP)}, 
  title={Correction of Automatic Speech Recognition with Transformer Sequence-To-Sequence Model}, 
  year={2020},
  volume={},
  number={},
  pages={7074-7078},
  doi={10.1109/ICASSP40776.2020.9053051}}

@article{DBLP:journals/corr/abs-2202-01157,
  author       = {Samrat Dutta and
                  Shreyansh Jain and
                  Ayush Maheshwari and
                  Ganesh Ramakrishnan and
                  Preethi Jyothi},
  title        = {Error Correction in {ASR} using Sequence-to-Sequence Models},
  journal      = {CoRR},
  volume       = {abs/2202.01157},
  year         = {2022},
  url          = {https://arxiv.org/abs/2202.01157},
  eprinttype    = {arXiv},
  eprint       = {2202.01157},
  bibsource    = {dblp computer science bibliography, https://dblp.org}
}

@inproceedings{ma23e_interspeech,
  title     = {{N-best T5: Robust ASR Error Correction using Multiple Input Hypotheses and Constrained Decoding Space}},
  author    = {Rao Ma and Mark J. F. Gales and Kate M. Knill and Mengjie Qian},
  year      = {2023},
  booktitle = {{Interspeech 2023}},
  pages     = {3267--3271},
  doi       = {10.21437/Interspeech.2023-1616},
  issn      = {2958-1796},
}

@misc{ma2023generativelargelanguagemodels,
      title={Can Generative Large Language Models Perform ASR Error Correction?}, 
      author={Rao Ma and Mengjie Qian and Potsawee Manakul and Mark Gales and Kate Knill},
      year={2023},
      eprint={2307.04172},
      archivePrefix={arXiv},
      primaryClass={cs.CL},
      url={https://arxiv.org/abs/2307.04172}, 
}

@ARTICLE{10930744,
  author={Ma, Rao and Qian, Mengjie and Gales, Mark and Knill, Kate},
  journal={IEEE Transactions on Audio, Speech and Language Processing}, 
  title={ASR Error Correction Using Large Language Models}, 
  year={2025},
  volume={33},
  number={},
  pages={1389-1401},
  doi={10.1109/TASLPRO.2025.3551083}}

@inproceedings{li24h_interspeech,
  title     = {{Investigating ASR Error Correction with Large Language Model and Multilingual 1-best Hypotheses}},
  author    = {Sheng Li and Chen Chen and Chin Yuen Kwok and Chenhui Chu and Eng Siong Chng and Hisashi Kawai},
  year      = {2024},
  booktitle = {{Interspeech 2024}},
  pages     = {1315--1319},
  doi       = {10.21437/Interspeech.2024-368},
  issn      = {2958-1796},
}

@misc{pu2024multistagelargelanguagemodel,
      title={Multi-stage Large Language Model Correction for Speech Recognition}, 
      author={Jie Pu and Thai-Son Nguyen and Sebastian Stüker},
      year={2024},
      eprint={2310.11532},
      archivePrefix={arXiv},
      primaryClass={cs.CL},
      url={https://arxiv.org/abs/2310.11532}, 
}

@inproceedings{yeen23_interspeech,
  title     = {{I Learned Error, I Can Fix It! : A Detector-Corrector Structure for ASR Error Calibration}},
  author    = {Heui-Yeen Yeen and Min-Ju Kim and Myoung-Wan Koo},
  year      = {2023},
  booktitle = {{Interspeech 2023}},
  pages     = {2693--2697},
  doi       = {10.21437/Interspeech.2023-2475},
  issn      = {2958-1796},
}

@inproceedings{malmi-etal-2019-encode,
    title = "Encode, Tag, Realize: High-Precision Text Editing",
    author = "Malmi, Eric  and
      Krause, Sebastian  and
      Rothe, Sascha  and
      Mirylenka, Daniil  and
      Severyn, Aliaksei",
    editor = "Inui, Kentaro  and
      Jiang, Jing  and
      Ng, Vincent  and
      Wan, Xiaojun",
    booktitle = "Proceedings of the 2019 Conference on Empirical Methods in Natural Language Processing and the 9th International Joint Conference on Natural Language Processing (EMNLP-IJCNLP)",
    month = nov,
    year = "2019",
    address = "Hong Kong, China",
    publisher = "Association for Computational Linguistics",
    url = "https://aclanthology.org/D19-1510/",
    doi = "10.18653/v1/D19-1510",
    pages = "5054--5065"
}

@inproceedings{stahlberg-kumar-2020-seq2edits,
    title = "{S}eq2{E}dits: Sequence Transduction Using Span-level Edit Operations",
    author = "Stahlberg, Felix  and
      Kumar, Shankar",
    editor = "Webber, Bonnie  and
      Cohn, Trevor  and
      He, Yulan  and
      Liu, Yang",
    booktitle = "Proceedings of the 2020 Conference on Empirical Methods in Natural Language Processing (EMNLP)",
    month = nov,
    year = "2020",
    address = "Online",
    publisher = "Association for Computational Linguistics",
    url = "https://aclanthology.org/2020.emnlp-main.418/",
    doi = "10.18653/v1/2020.emnlp-main.418",
    pages = "5147--5159"
}

@inproceedings{10.1609/aaai.v37i11.26531,
author = {Leng, Yichong and Tan, Xu and Liu, Wenjie and Song, Kaitao and Wang, Rui and Li, Xiang-Yang and Qin, Tao and Lin, Ed and Liu, Tie-Yan},
title = {SoftCorrect: error correction with soft detection for automatic speech recognition},
year = {2023},
isbn = {978-1-57735-880-0},
publisher = {AAAI Press},
url = {https://doi.org/10.1609/aaai.v37i11.26531},
doi = {10.1609/aaai.v37i11.26531},
booktitle = {Proceedings of the Thirty-Seventh AAAI Conference on Artificial Intelligence and Thirty-Fifth Conference on Innovative Applications of Artificial Intelligence and Thirteenth Symposium on Educational Advances in Artificial Intelligence},
articleno = {1462},
numpages = {9},
series = {AAAI'23/IAAI'23/EAAI'23}
}

@Article{app10196936,
AUTHOR = {Bang, Jeong-Uk and Yun, Seung and Kim, Seung-Hi and Choi, Mu-Yeol and Lee, Min-Kyu and Kim, Yeo-Jeong and Kim, Dong-Hyun and Park, Jun and Lee, Young-Jik and Kim, Sang-Hun},
TITLE = {KsponSpeech: Korean Spontaneous Speech Corpus for Automatic Speech Recognition},
JOURNAL = {Applied Sciences},
VOLUME = {10},
YEAR = {2020},
NUMBER = {19},
ARTICLE-NUMBER = {6936},
URL = {https://www.mdpi.com/2076-3417/10/19/6936},
ISSN = {2076-3417},
DOI = {10.3390/app10196936}
}

@inproceedings{ha20_interspeech,
  title     = {{ClovaCall: Korean Goal-Oriented Dialog Speech Corpus for Automatic Speech Recognition of Contact Centers}},
  author    = {Jung-Woo Ha and Kihyun Nam and Jingu Kang and Sang-Woo Lee and Sohee Yang and Hyunhoon Jung and Hyeji Kim and Eunmi Kim and Soojin Kim and Hyun Ah Kim and Kyoungtae Doh and Chan Kyu Lee and Nako Sung and Sunghun Kim},
  year      = {2020},
  booktitle = {{Interspeech 2020}},
  pages     = {409--413},
  doi       = {10.21437/Interspeech.2020-1136},
  issn      = {2958-1796},
}

@inproceedings{koo-etal-2023-kebap,
    title = "{KEBAP}: {K}orean Error Explainable Benchmark Dataset for {ASR} and Post-processing",
    author = "Koo, Seonmin  and
      Park, Chanjun  and
      Kim, Jinsung  and
      Seo, Jaehyung  and
      Eo, Sugyeong  and
      Moon, Hyeonseok  and
      Lim, Heuiseok",
    editor = "Bouamor, Houda  and
      Pino, Juan  and
      Bali, Kalika",
    booktitle = "Proceedings of the 2023 Conference on Empirical Methods in Natural Language Processing",
    month = dec,
    year = "2023",
    address = "Singapore",
    publisher = "Association for Computational Linguistics",
    url = "https://aclanthology.org/2023.emnlp-main.292/",
    doi = "10.18653/v1/2023.emnlp-main.292",
    pages = "4798--4815"
}

@inproceedings{park-etal-2024-hyper,
    title = "Hyper-{BTS} Dataset: Scalability and Enhanced Analysis of Back {T}ran{S}cription ({BTS}) for {ASR} Post-Processing",
    author = "Park, Chanjun  and
      Seo, Jaehyung  and
      Lee, Seolhwa  and
      Son, Junyoung  and
      Moon, Hyeonseok  and
      Eo, Sugyeong  and
      Lee, Chanhee  and
      Lim, Heuiseok",
    editor = "Graham, Yvette  and
      Purver, Matthew",
    booktitle = "Findings of the Association for Computational Linguistics: EACL 2024",
    month = mar,
    year = "2024",
    address = "St. Julian{'}s, Malta",
    publisher = "Association for Computational Linguistics",
    url = "https://aclanthology.org/2024.findings-eacl.5/",
    doi = "10.18653/v1/2024.findings-eacl.5",
    pages = "67--78"
}

@inproceedings{10.1145/3394486.3406703,
author = {Rasley, Jeff and Rajbhandari, Samyam and Ruwase, Olatunji and He, Yuxiong},
title = {DeepSpeed: System Optimizations Enable Training Deep Learning Models with Over 100 Billion Parameters},
year = {2020},
isbn = {9781450379984},
publisher = {Association for Computing Machinery},
address = {New York, NY, USA},
url = {https://doi.org/10.1145/3394486.3406703},
doi = {10.1145/3394486.3406703},
booktitle = {Proceedings of the 26th ACM SIGKDD International Conference on Knowledge Discovery \& Data Mining},
pages = {3505–3506},
numpages = {2},
location = {Virtual Event, CA, USA},
series = {KDD '20}
}

@inproceedings{meripo22_interspeech,
  title     = {{ASR Error Detection via Audio-Transcript entailment}},
  author    = {Nimshi Venkat Meripo and Sandeep Konam},
  year      = {2022},
  booktitle = {{Interspeech 2022}},
  pages     = {3358--3362},
  doi       = {10.21437/Interspeech.2022-11177},
  issn      = {2958-1796},
}

@inproceedings{devlin-etal-2019-bert,
    title = "{BERT}: Pre-training of Deep Bidirectional Transformers for Language Understanding",
    author = "Devlin, Jacob  and
      Chang, Ming-Wei  and
      Lee, Kenton  and
      Toutanova, Kristina",
    editor = "Burstein, Jill  and
      Doran, Christy  and
      Solorio, Thamar",
    booktitle = "Proceedings of the 2019 Conference of the North {A}merican Chapter of the Association for Computational Linguistics: Human Language Technologies, Volume 1 (Long and Short Papers)",
    month = jun,
    year = "2019",
    address = "Minneapolis, Minnesota",
    publisher = "Association for Computational Linguistics",
    url = "https://aclanthology.org/N19-1423/",
    doi = "10.18653/v1/N19-1423",
    pages = "4171--4186"
}

@INPROCEEDINGS{10448230,
  author={Harvill, John and Khaziev, Rinat and Li, Scarlett and Cogill, Randy and Wang, Lidan and Chennupati, Gopinath and Thadakamalla, Hari},
  booktitle={ICASSP 2024 - 2024 IEEE International Conference on Acoustics, Speech and Signal Processing (ICASSP)}, 
  title={Significant ASR Error Detection for Conversational Voice Assistants}, 
  year={2024},
  volume={},
  number={},
  pages={11606-11610},
  doi={10.1109/ICASSP48485.2024.10448230}}

@inproceedings{gekhman-etal-2022-red,
    title = "{RED}-{ACE}: Robust Error Detection for {ASR} using Confidence Embeddings",
    author = "Gekhman, Zorik  and
      Zverinski, Dina  and
      Mallinson, Jonathan  and
      Beryozkin, Genady",
    editor = "Goldberg, Yoav  and
      Kozareva, Zornitsa  and
      Zhang, Yue",
    booktitle = "Proceedings of the 2022 Conference on Empirical Methods in Natural Language Processing",
    month = dec,
    year = "2022",
    address = "Abu Dhabi, United Arab Emirates",
    publisher = "Association for Computational Linguistics",
    url = "https://aclanthology.org/2022.emnlp-main.180/",
    doi = "10.18653/v1/2022.emnlp-main.180",
    pages = "2800--2808"
}

@article{anshor2025implementing,
  title={Implementing type-2 fuzzy logic for post-ASR correction in low-resource languages: A case study in Sundanese},
  author={Anshor, Abdul Halim and Wiyatno, Tri Ngudi},
  journal={Uncertainty Discourse and Applications},
  volume={2},
  number={3},
  pages={196--204},
  year={2025}
}

@article{10.1145/3793254,
author = {Bhandari, Abhishek and Harit, Gaurav},
title = {Post-ASR Correction for Low-Resource Rajasthani Language},
year = {2026},
publisher = {Association for Computing Machinery},
address = {New York, NY, USA},
issn = {2375-4699},
url = {https://doi.org/10.1145/3793254},
doi = {10.1145/3793254},
note = {Just Accepted},
journal = {ACM Trans. Asian Low-Resour. Lang. Inf. Process.},
month = jan
}

@inproceedings{znotins-gruzitis-2025-conversational,
    title = "From Conversational Speech to Readable Text: Post-Processing Noisy Transcripts in a Low-Resource Setting",
    author = "Znotins, Arturs  and
      Gruzitis, Normunds  and
      Dargis, Roberts",
    editor = "Bak, JinYeong  and
      Goot, Rob van der  and
      Jang, Hyeju  and
      Buaphet, Weerayut  and
      Ramponi, Alan  and
      Xu, Wei  and
      Ritter, Alan",
    booktitle = "Proceedings of the Tenth Workshop on Noisy and User-generated Text",
    month = may,
    year = "2025",
    address = "Albuquerque, New Mexico, USA",
    publisher = "Association for Computational Linguistics",
    url = "https://aclanthology.org/2025.wnut-1.15/",
    doi = "10.18653/v1/2025.wnut-1.15",
    pages = "143--148",
    ISBN = "979-8-89176-232-9"
}

@inproceedings{aycock2025can,
  title={Can {LLM}s Really Learn to Translate a Low-Resource Language from One Grammar Book?},
  author={Seth Aycock and David Stap and Di Wu and Christof Monz and Khalil Sima'an},
  booktitle={The Thirteenth International Conference on Learning Representations},
  year={2025},
  url={https://openreview.net/forum?id=aMBSY2ebPw}
}

@inproceedings{cahyawijaya-etal-2024-llms,
    title = "{LLM}s Are Few-Shot In-Context Low-Resource Language Learners",
    author = "Cahyawijaya, Samuel  and
      Lovenia, Holy  and
      Fung, Pascale",
    editor = "Duh, Kevin  and
      Gomez, Helena  and
      Bethard, Steven",
    booktitle = "Proceedings of the 2024 Conference of the North American Chapter of the Association for Computational Linguistics: Human Language Technologies (Volume 1: Long Papers)",
    month = jun,
    year = "2024",
    address = "Mexico City, Mexico",
    publisher = "Association for Computational Linguistics",
    url = "https://aclanthology.org/2024.naacl-long.24/",
    doi = "10.18653/v1/2024.naacl-long.24",
    pages = "405--433"
}

@phdthesis{6ff6bbca842f4a909c6ea0d3cce748da,
title = "Measuring Dialect Pronunciation Differences using Levenshtein Distance",
author = "Heeringa, \{Wilbert Jan\}",
note = "Relation: http://www.rug.nl/ date\_submitted:2004 Rights: University of Groningen",
year = "2004",
language = "English",
publisher = "s.n.",
school = "University of Groningen",
}

@misc{park2020koelectra,
  author = {Park, Jangwon},
  title = {KoELECTRA: Pretrained ELECTRA Model for Korean},
  year = {2020},
  publisher = {GitHub},
  journal = {GitHub repository},
  howpublished = {\url{https://github.com/monologg/KoELECTRA}}
}

@inproceedings{
Clark2020ELECTRA,
title={ELECTRA: Pre-training Text Encoders as Discriminators Rather Than Generators},
author={Kevin Clark and Minh-Thang Luong and Quoc V. Le and Christopher D. Manning},
booktitle={International Conference on Learning Representations},
year={2020},
url={https://openreview.net/forum?id=r1xMH1BtvB}
}

@inproceedings{xue-etal-2021-mt5,
    title = "m{T}5: A Massively Multilingual Pre-trained Text-to-Text Transformer",
    author = "Xue, Linting  and
      Constant, Noah  and
      Roberts, Adam  and
      Kale, Mihir  and
      Al-Rfou, Rami  and
      Siddhant, Aditya  and
      Barua, Aditya  and
      Raffel, Colin",
    booktitle = "Proceedings of the 2021 Conference of the North American Chapter of the Association for Computational Linguistics: Human Language Technologies",
    month = jun,
    year = "2021",
    address = "Online",
    publisher = "Association for Computational Linguistics",
    url = "https://aclanthology.org/2021.naacl-main.41",
    doi = "10.18653/v1/2021.naacl-main.41",
    pages = "483--498"
}

@software{paust_pkot5_v1,
  author = {Dennis Park},
  month = {5},
  title = {pko-t5: PAUST Korean T5 for text-to-text unified framework},
  url = {https://github.com/paust-team/pko-t5},
  version = {1.0},
  year = {2022}
}

@misc{grattafiori2024llama3herdmodels,
  title={The llama 3 herd of models},
  author={Grattafiori, Aaron and Dubey, Abhimanyu and Jauhri, Abhinav and Pandey, Abhinav and Kadian, Abhishek and Al-Dahle, Ahmad and Letman, Aiesha and Mathur, Akhil and Schelten, Alan and Vaughan, Alex and others},
  year={2024},
  eprint={2407.21783},
  archivePrefix={arXiv},
  url={https://arxiv.org/abs/2407.21783},
}

@misc{qwen2025qwen25technicalreport,
      title={Qwen2.5 Technical Report}, 
      author={Qwen and An Yang and Baosong Yang and Beichen Zhang and Binyuan Hui and Bo Zheng and Bowen Yu and Chengyuan Li and Dayiheng Liu and Fei Huang and Haoran Wei and Huan Lin and Jian Yang and Jianhong Tu and Jianwei Zhang and Jianxin Yang and Jiaxi Yang and Jingren Zhou and Junyang Lin and Kai Dang and Keming Lu and Keqin Bao and Kexin Yang and Le Yu and Mei Li and Mingfeng Xue and Pei Zhang and Qin Zhu and Rui Men and Runji Lin and Tianhao Li and Tianyi Tang and Tingyu Xia and Xingzhang Ren and Xuancheng Ren and Yang Fan and Yang Su and Yichang Zhang and Yu Wan and Yuqiong Liu and Zeyu Cui and Zhenru Zhang and Zihan Qiu},
      year={2025},
      eprint={2412.15115},
      archivePrefix={arXiv},
      primaryClass={cs.CL},
      url={https://arxiv.org/abs/2412.15115}, 
}

@inproceedings{
loshchilov2018decoupled,
title={Decoupled Weight Decay Regularization},
author={Ilya Loshchilov and Frank Hutter},
booktitle={International Conference on Learning Representations},
year={2019},
url={https://openreview.net/forum?id=Bkg6RiCqY7},
}

@inproceedings{
hu2022lora,
title={Lo{RA}: Low-Rank Adaptation of Large Language Models},
author={Edward J Hu and yelong shen and Phillip Wallis and Zeyuan Allen-Zhu and Yuanzhi Li and Shean Wang and Lu Wang and Weizhu Chen},
booktitle={International Conference on Learning Representations},
year={2022},
url={https://openreview.net/forum?id=nZeVKeeFYf9}
}

@inproceedings{NEURIPS_DATASETS_AND_BENCHMARKS_2021_98dce83d,
 author = {Park, Sungjoon and Moon, Jihyung and Kim, Sungdong and Cho, Won Ik and Han, Ji Yoon and Park, Jangwon and Song, Chisung and Kim, Junseong and Song, Youngsook and Oh, Taehwan and Lee, Joohong and Oh, Juhyun and Lyu, Sungwon and Jeong, Younghoon and Lee, Inkwon and Seo, Sangwoo and Lee, Dongjun and Kim, Hyunwoo and Lee, Myeonghwa and Jang, Seongbo and Do, Seungwon and Kim, Sunkyoung and Lim, Kyungtae and Lee, Jongwon and Park, Kyumin and Shin, Jamin and Kim, Seonghyun and Park, Lucy and Park, Lucy and Oh, Alice and Ha (NAVER AI Lab), Jung-Woo and Cho, Kyunghyun and Cho, Kyunghyun},
 booktitle = {Proceedings of the Neural Information Processing Systems Track on Datasets and Benchmarks},
 editor = {J. Vanschoren and S. Yeung},
 pages = {},
 title = {KLUE: Korean Language Understanding Evaluation},
 url = {https://datasets-benchmarks-proceedings.neurips.cc/paper_files/paper/2021/file/98dce83da57b0395e163467c9dae521b-Paper-round2.pdf},
 volume = {1},
 year = {2021}
}

@article{haghrah2025pyit2fls,
  title={PyIT2FLS: An open-source Python framework for flexible and scalable development of type 1 and interval type 2 fuzzy logic models},
  author={Haghrah, Amir Arslan and Ghaemi, Sehraneh and Badamchizadeh, Mohammad Ali},
  journal={SoftwareX},
  volume={30},
  pages={102146},
  year={2025},
  publisher={Elsevier}
}
